\def\paperlanguage{}
\pdfoutput=1 % for arxiv

\documentclass[letterpaper, 10 pt, conference]{ieeeconf}  % Comment this line out if you need a4paper

\usepackage{bm}
\usepackage{cite}
\usepackage{flushend}
\include{preamble}

\IEEEoverridecommandlockouts                              % This command is only needed if
\title{\LARGE \textbf
  {
    \switchlanguage%
    {%
      Tool Shape Optimization through Backpropagation of Neural Network
    }%
    {%
      ニューラルネットワークの誤差逆伝播による道具形状最適化
    }%
  }
}

\author{Kento Kawaharazuka$^1$, Toru Ogawa$^2$, and Cota Nabeshima$^2$% <-this % stops a space
  \thanks{$^1$ An author is associated with Department of Mechano-Informatics, Graduate School of Information Science and Technology, The University of Tokyo. %, 7-3-1 Hongo, Bunkyo-ku, Tokyo, 113-8656, Japan.
    \texttt\small kawaharazuka@jsk.t.u-tokyo.ac.jp
  }
  \thanks{$^2$ Authors are associated with Preferred Networks, Inc. %
    \texttt\small \{ogawa, cota\}@preferred.jp
  }
  \thanks{
  This work is an achievement during part-time job at Preferred Networks. %
  }
}
\begin{document}

\maketitle
\thispagestyle{empty}
\pagestyle{empty}

%%%%%%%%%%%%%%%%%%%%%%%%%%%%%%%%%%%%%%%%%%%%%%%%%%%%%%%%%%%%%%%%%%%%%%%%%%%%%%%%
\begin{abstract}
  \switchlanguage%
  {%
    When executing a certain task, human beings can choose or make an appropriate tool to achieve the task.
    This research especially addresses the optimization of tool shape for robotic tool-use.
    We propose a method in which a robot obtains an optimized tool shape, tool trajectory, or both, depending on a given task.
    The feature of our method is that a transition of the task state when the robot moves a certain tool along a certain trajectory is represented by a deep neural network.
    We applied this method to object manipulation tasks on a 2D plane, and verified that appropriate tool shapes are generated by using this novel method.
  }%
  {%
    人間はあるタスクを行う際に, それに適切な道具を選択, または製作し, 利用することができる.
    本研究はその中でも特に, タスクに応じた適切な道具の形状最適化について議論する.
    道具とその動作軌道によるタスク状態の遷移をニューラルネットワークにより記述する.
    あるタスクが与えられた際に, それに最適な道具形状・動作軌道, またはその両方を得ることができる手法を提案する.
    本手法を2次元平面上の物体移動タスク適用し, その有効性を確認した.
  }%
\end{abstract}

\section{INTRODUCTION}\label{sec:introduction}
\begin{figure}[t]
  \centering
  \includegraphics[width=0.75\columnwidth]{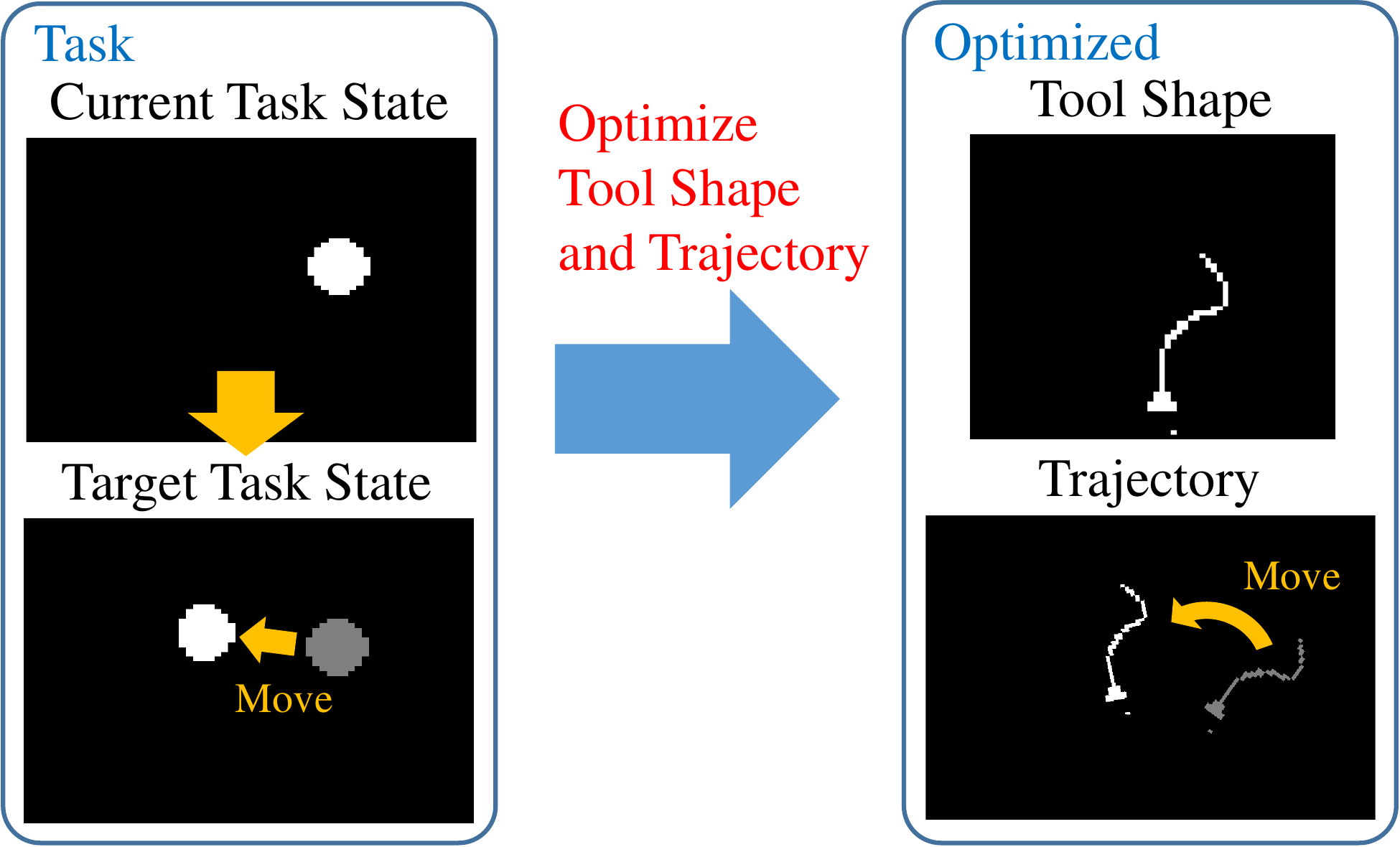}
  \caption{Functional diagram of our method where a tool shape and trajectory are optimized for a given task from a current and target task state}
  \label{figure:motivation}
  \vspace{-3.0ex}
\end{figure}

\switchlanguage%
{%
  Tool-use is one of the fundamental abilities of human beings.
  When executing a task, human beings can choose or make an appropriate tool to achieve the task.
  For a robot to work in a human environment, combining existing tools or making new tools is necessary to expand its capability.
  In this study, we mainly focus on the optimization of tool shape and tool trajectory, as a foundation of tool making.

  Robotic tool-use has been studied in various topics: tool recognition \cite{huber2004template, kemp2006tooltip}, tool understanding \cite{zhu2015understanding, myers2015affordance}, tool choice \cite{saito2018tool, tee2018tool}, and motion generation with tool-use \cite{okada2006tool, toussaint2018tooluse, fang2018toolgrasp}.
  However, there have been few studies about tool-making or making a new appropriate tool for a given task.
  Nair, et al. developed methods to construct a new tool by combining two existing tools using geometric reasoning \cite{nair2019macgyvering, nair2019tool}.
  Wicaksono, et al. developed frameworks of tool creation as an extension of tool-use learning \cite{wicaksono2015tool, wicaksono2017tool}.
  However, because \cite{nair2019macgyvering, nair2019tool} can generate only tools expressed by the combination of two existing tools and \cite{wicaksono2015tool, wicaksono2017tool} can generate only tools similar to a reference tool due to random generation of tools fulfilling many hypotheses (e.g. a hook-like tool), various free forms of tool shapes cannot be handled.
  Also, because \cite{wicaksono2015tool, wicaksono2017tool} must be tested by the actual robot to choose the best tool, the optimization of tool shape takes too much time.

  Apart from robotic tool-use, an optimization method of robot design parameters such as link length and actuator placement has been developed \cite{ha2017optimization}.
  Also, some studies have jointly optimized the robot design parameters and control scheme using a genetic algorithm \cite{sims1994evolving, lipson2000evolving} or reinforcement learning \cite{schaff2019jointly}, in simulation environment.
  While these studies are similar to the scenario of tool making, there are two problems in common.
  First, these robot designs (e.g. link length and actuator placement) are manually parameterized and various free forms cannot be handled.
  To appropriately parameterize the design, prior knowledge of human experts is necessary.
  Second, experiments of almost all previous works are conducted in simulation environment.
  This is because we must move the robot to obtain evaluation value in the process of optimization and it takes too much time in the actual environment.
  Also, these studies cannot consider the characteristics of the actual environment, e.g. friction, hysteresis, and robot model error.
  % Huber, et al. developed a tool recognition method using template matching \cite{huber2004template}, and Kemp, et al. developed tool tip recognition and its control system \cite{kemp2006tooltip}.
  % Zhu, et al. extracted tool function and motion trajectories from videos of human tool-use \cite{zhu2015understanding}, and Myers, et al. extracted tool affordance using deep learning \cite{myers2015affordance}.
  % Saito, et al. developed a learning system of tool choice from tool-use experience \cite{saito2018tool}, and Tee, et al. developed a tool choice system using the analogy of the robot body and tool shape \cite{tee2018tool}.
  % Okada, et al. extracted essential elements to describe tool-use by programming \cite{okada2006tool}.
  % Taussaint, et al. studied motion planning for dynamic tool-use \cite{toussaint2018tooluse}.
  % Nabeshima, et al. suggested the relationship between the adaptation of body representation and tool-use in the context of robotics \cite{nabeshima2006tool}.
  % Fang, et al. explored how to hold a tool depending on the task \cite{fang2018toolgrasp}.

  % By studying this topic, a robot is expected to become able to use a tool even if it is deformable like a metal wire.
  % As a further extension, the robot would become able to combine various existing tools designed for humans and make other tools by tools.
  Our contributions of this study are as below,
  \begin{itemize}
    \item
      We use an image to represent a tool shape.
      A tool shape can be easily converted to an image, and we can uniformly handle various tool shapes without prior knowledge.
    \item
      All experiments including evaluations are conducted in the actual environment.
      By acquiring a tool-use model using the actual robot data of random movements, the tool shape is directly optimized without evaluating the movements of the actual robot.
  \end{itemize}

  As shown in \figref{figure:motivation}, this study simultaneously calculates an optimized tool shape and trajectory for a given task.
  A transition of the task state when a robot moves a certain tool along a certain trajectory is represented by a deep neural network.
  An optimized tool shape, tool trajectory, or both for a target task, can be obtained by using the backpropagation technique \cite{rumelhart1986backprop} of the neural network.
  Although this method includes motion trajectory optimization, we mainly put stress on tool shape optimization.
  % Even though a tool shape is potentially complex and is large in variety for a task, our novel optimization method can generate an appropriate tool shape for the task.
  We conduct experiments using the actual robot on a 2D plane to verify the effectiveness of this study.

  % In the following sections, firstly in \secref{sec:tool-net}, we will explain the details of Tool Shape and Trajectory Optimization Network (Tool-Net), and optimization method using the network.
  % In \secref{sec:tool-net}, we will conduct experiments using the robot moving on a 2D plane, and verify the effectiveness of this study.
}%
{%
  道具利用は人間の本質的な能力の一つである.
  人間はあるタスクを行う際に, それに適切な道具を選択, または製作し, 適切な動作軌道で利用することができる.
  ロボットが人間と同じように作業するためには, 既存の道具を組み合わせたり, 状況に合わせた新しい道具を製作したりすることが必要である.
  本研究ではその基盤となる技術である, 道具の最適化に着目する.

  これまで, ロボットにおける道具使用に必要な要素である, 道具認識, 道具理解, 道具選択, 道具を用いた動作生成に関して様々な研究が行われている.
  Huberらはテンプレートマッチングにより道具の認識を行い\cite{huber2004template}, Kempらは道具の先端の認識とその制御手法を開発した\cite{kemp2006tooltip}.
  Zhuらは人間の道具使用動画から道具の機能や動作軌道の抽出を行い\cite{zhu2015understanding}, Myersらは道具のaffordanceを深層学習により抽出した\cite{myers2015affordance}.
  Saitoらはロボットの道具使用経験から道具選択を学習する枠組みを作り\cite{saito2018tool}, Teeらは自分の身体と道具の類似性から道具選択を行う仕組みを開発した\cite{tee2018tool}.
  Okadaらは道具使用の記述に必要な要素の抽出と視覚による確認\cite{okada2006tool}, Taussaintらはダイナミックな道具利用とその動作計画\cite{toussaint2018tooluse}について研究している.
  Nabeshimaらは道具による身体図式の変容について考察し\cite{nabeshima2006tool}, Fangらはタスクに応じた道具の持ち方を選択する手法を開発した\cite{fang2018toolgrasp}.
  しかし, タスクに応じた新たな道具生成に関する研究はこれまでにない.

  そこで本研究では, 道具の形状最適化とその利用について議論する.
  つまり, \figref{figure:motivation}のように, あるタスクが与えられたときに, そのタスクに適した道具の形状と軌道を出力するという問題である.
  この問題を解くことにより, 針金のように変形する道具が扱えるようになる.
  またその発展形として, 道具を組み合わせたり, 実際に人間のように工具を使って道具を製作することができるようになると考える.

  本研究では, 道具とその動作軌道によるタスク状態の遷移をニューラルネットワークにより表現する.
  そして, あるタスクが与えられた際に, ニューラルネットワークの誤差逆伝播法用いることで, そのタスクに最適な道具形状・動作軌道, またはその両方を得ることができる.
  本手法は道具の動作軌道計画も含むが, 主に道具の形状最適化について議論をする.
  先行研究として, 道具ではないが, ロボット設計におけるリンク長・アクチュエータ配置の身体パラメータの最適化研究も存在する\cite{ha2017optimization}.
  % 道具の形はより複雑で多様性があり, 本研究は道具を画像という自由度の高い表現形式によって表し, それを変化させていくという点でこれらの研究とは異なる.
  道具の形はより複雑で多様性があり, 本研究はそれらを最適化していく手法について考える.

  以降では, まず第二章で, 本研究で提案する道具形状・動作軌道最適化ネットワーク(Tool-Net)の詳細と, それを用いた道具形状・動作軌道最適化について述べる.
  第三章では, 平面状を動く2自由度ロボットにおいて実験を行い, 本手法の有効性を確認する.
  最後に, 本研究に関する議論と結論を述べる.
}%

\section{Tool Shape and Trajectory Optimization Network} \label{sec:tool-net}
\switchlanguage%
{%
  In this study, we represent a transition of task state, when using a given tool shape and trajectory, by a neural network.
  We call this network ``Tool Shape and Trajectory Optimization Network (Tool-Net)''.

 In the following sections, we assume that the robot moves on a 2D plane and the task is an object manipulation task, for simplicity.
 In \secref{sec:discussion}, we will discuss extensions of our method to a robot that moves in 3D space and other kinds of tasks.
}%
{%
  本研究では, 与えられた道具形状とその動作軌道によるタスク状態の遷移をニューラルネットワークにより表現し用いる.
  このネットワークをTool-Net (Tool Shape and Trajectory Optimization Network)と呼ぶ.

  本章ではロボットの動きを2次元平面上に限定して実装を考えるが, 本手法は3次元に拡張することも可能であり, これについては\secref{sec:discussion}にて議論する.
  また, 本研究では\figref{figure:motivation}に示すように道具による物体の移動タスクについて考えるが, 本研究の手法の適用先はこれに限らず, これについても\secref{sec:discussion}にて議論する.
}%

\subsection{Network Structure of Tool-Net} \label{subsec:basic-structure}
\begin{figure}[t]
  \centering
  \includegraphics[width=1.0\columnwidth]{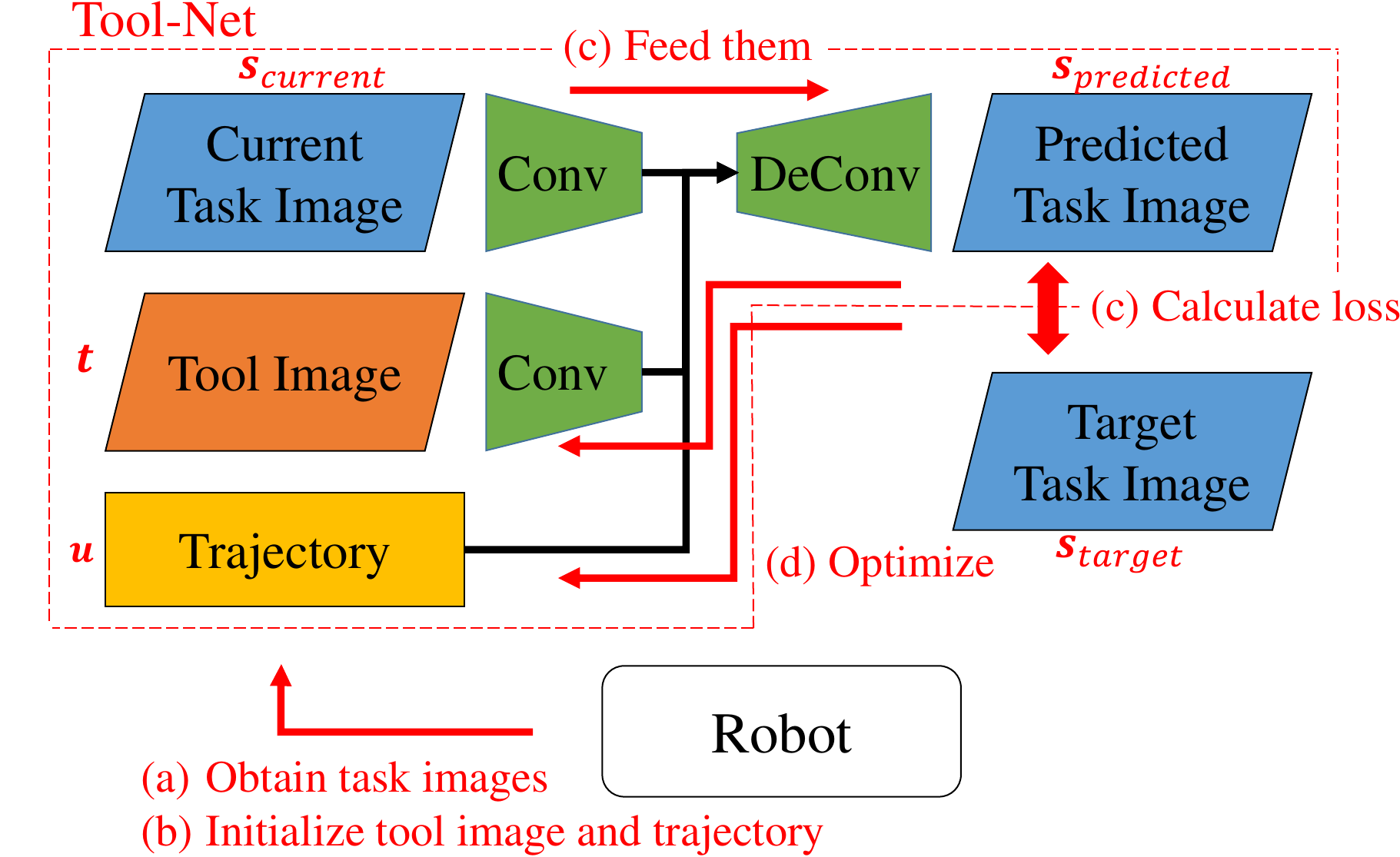}
  \vspace{-3.0ex}
  \caption{Network structure of the proposed Tool-Net and optimization procedure of tool shape and trajectory}
  \label{figure:network-structure}
  \vspace{-3.0ex}
\end{figure}

\switchlanguage%
{%
  The network structure of Tool-Net is shown in \figref{figure:network-structure}.
  This network is represented by the equation below,
  \begin{align}
    \bm{s}_{predicted} = \bm{f}(\bm{s}_{current}, \bm{t}, \bm{u}) \label{eq:tool-net}
  \end{align}
  where $\bm{s}_{current}$ is a current task state, $\bm{t}$ is a tool shape, $\bm{u}$ is a tool trajectory, $\bm{s}_{predicted}$ is a predicted task state after moving the robot using $\bm{u}$ and $\bm{t}$, and $\bm{f}$ represents Tool-Net.
  $\bm{f}$ is trained, and $\bm{t}$ and $\bm{u}$ are optimized when given a target task state $\bm{s}_{target}$.
  %Various representations of $\bm{s}$ and $\bm{t}$ can be considered, and especially the representation of $\bm{s}$ largely depends on the task.
  Because we handle object manipulation tasks in this study, we use a binarized image as shown in \figref{figure:motivation}, which can flexibly express the object position and posture, as task state $\bm{s}$.
  We also use a binarized image, which has high degrees of freedom, as tool image $\bm{t}$.
  % Although the representations of $\bm{t}$ such as parameters of spline curve and combination parameters of tool primitives can be considered, they are not suitable for generating new tool shapes, because human prior knowledge limits ideas for tool shape.
  In binarized images, the background color is black (its value is 0), and the tool and manipulated object color is white (its value is 1).
  Regarding tool trajectory, we assume quasi-static movement and constant joint velocity in this study, and so we represent $\bm{u}$ as $(\bm{\theta}^{T}_{start}, \bm{\theta}^{T}_{end})^{T}$.
  $\bm{\theta}_{\{start, end\}}$ is the starting or ending joint angles of the robot, and the whole trajectory is the trajectory that interpolates these joint angles linearly by constant velocity.

  In detail, $\bm{s}_{current}$ and $\bm{t}$ are inputted through convolutional layers and concatenated with $\bm{u}$, and $\bm{s}_{predicted}$ is outputted through deconvolutional layers.
}%
{%
  まず, Tool-Netのネットワーク構造を\figref{figure:network-structure}に示す.
  これは式で表すと, 以下の式と同等である.
  \begin{align}
    \bm{s}_{predicted} = \bm{f}(\bm{s}_{current}, \bm{t}, \bm{u}) \label{eq:tool-net}
  \end{align}
  ここで, $\bm{s}_{current}$は現在タスク状態, $\bm{t}$は道具形状, $\bm{u}$は動作軌道, $\bm{s}_{predicted}$はロボット動作後に予測されるタスク状態, $\bm{f}$はTool-Netを表す.
  本研究では, $\bm{f}$を学習し, 指令タスク状態$\bm{s}_{target}$を与えることで$\bm{t}, \bm{u}$を最適化していくことになる.
  このタスク状態$\bm{s}$, 道具形状$\bm{t}$はどのように表現をしても良く, 特にタスク状態は扱うタスクによって表現が大きく異なる.
  本研究では物体移動タスクを扱うため, 物体の位置姿勢を柔軟に表現可能な, \figref{figure:motivation}に示したような二値画像をタスク状態として用いる.
  道具の表現にはスプライン曲線のパラメータや道具の道具プリミティブの組み合わせのパラメータ等が考えられるが, それは道具の表現に事前知識を利用することになり, 新たな道具生成には向かない.
  よって, 本研究ではその自由度の高さから道具も\figref{figure:motivation}に示したような二値画像で表すこととする.
  ２値画像においては, 背景が黒(0), 道具や操作物体が白(1)とする.
  また, 動作軌道$\bm{u}$にも様々な表現があるが, 本研究では簡単のため準静的動作かつ動作速度一定を仮定し, $\bm{u}=(\bm{\theta}^{T}_{start}, \bm{\theta}^{T}_{end})^{T}$としている.
  ここで, $\bm{\theta}_{start}$, $\bm{\theta}_{end}$は動作軌道の最初と最後の関節角度を表し, これを準静的に一定速度で線形に補間したものが動作軌道となる.

  実際のニューラルネットワークでは, $\bm{s}_{current}$と$\bm{t}$を畳み込み層を通して, $\bm{u}$と結合し, それを逆畳み込みによって$\bm{s}_{predicted}$を出力するようなネットワークとなる.
}%

\subsection{Data Collection for Tool-Net} \label{subsec:data-collection}
\switchlanguage%
{%
  The procedures to collect data for training of Tool-Net are as below.
  \renewcommand{\labelenumi}{(\alph{enumi})}
  \begin{enumerate}
    \item Initialize the robot posture and attach a randomly generated tool to the manipulator tip
    \item Obtain the tool shape image $\bm{t}$
    \item Set $\bm{\theta}_{start}$ randomly and move the robot
    \item Randomly place an object
    \item Obtain the task state image $\bm{s}_{start}$
    \item Set $\bm{\theta}_{end}$ randomly and move the robot
    \item Obtain the task state image $\bm{s}_{end}$
    \item Repeat (c) -- (g), collect the data, and go back to (a) after the conditions (described below) are satisfied
  \end{enumerate}

\begin{figure}[t]
  \centering
  \includegraphics[width=0.8\columnwidth]{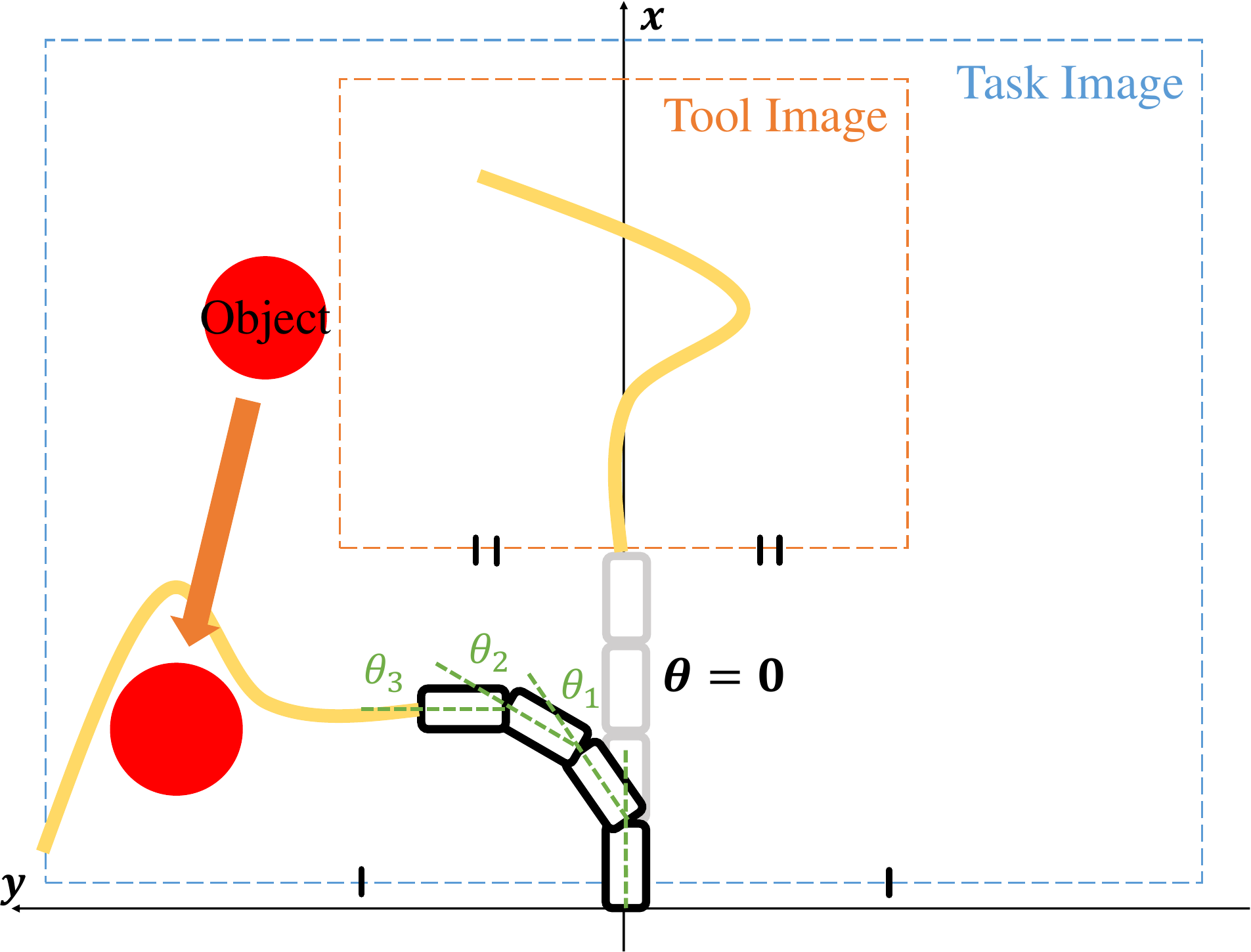}
  \caption{The settings of the robot, task state, and tool state}
  \label{figure:robot-config}
  \vspace{-3.0ex}
\end{figure}

% \begin{figure}[t]
%   \centering
%   \includegraphics[width=1.0\columnwidth]{figs/random-tool}
%   \caption{The examples of generated tool shapes by randomly changing control points of 3-dimensional B-spline (tool lengths are adjusted).}
%   \label{figure:random-tool}
%   \vspace{-1.0zh}
% \end{figure}

  In (a), the joint angle of the robot is initialized to $\bm{0}$.
  We define the joint angles as shown in \figref{figure:robot-config}, and so the initialized posture is straightly aligned.
  % Also, we need to prepare various tool shapes to attach to the robot.
  % Because we use a metal wire as the tool, which will be explained in \secref{subsec:experimental-setup}, we generate cubic b-splines whose control points are randomly arranged and whose lengths are aligned to constant lengths as shown in \figref{figure:random-tool}, and we make tools based on these shapes by hand.
  % Because we use a metal wire as the tool, which will be explained in \secref{subsec:experimental-setup},  we make tools randomly by hand.
  % Tools may be made by 3D printer or clay, and not limited to metal wire.
  To prepare various tool shapes to be attached to the robot, we make tools randomly by hand with a metal wire, which will be explained in \secref{subsec:experimental-setup}.
  3D printing or clay modeling are also applicable for the purpose.

  In (c) and (f), when the robot is randomly moved, the limit of $\bm{\theta}_{min} \leq \bm{\theta}_{\{start, end\}} \leq \bm{\theta}_{max}$ is set ($\bm{\theta}_{\{min, max\}}$ is the lower or upper limit of joint angles).
  Also, because the robot hardly moves when $\bm{\theta}_{end}$ is close to $\bm{\theta}_{start}$, we set the limit of $||\bm{\theta}_{start}-\bm{\theta}_{end}||_{1}>\theta_{thre}$ when randomly choosing the $\bm{\theta}_{\{start, end\}}$ ($||\cdot||_{1}$ expresses L1 norm).

  In (h), we use a sampling technique to balance changed samples and unchanged samples.
  A changed sample means that the robot moves the target object, and an unchanged sample means that the robot does not touch the target object.
  Due to the random trajectory, we get more unchanged samples than changed samples.
  We use symmetric chamfer distance $d_{chamfer}$ \cite{borgefors1988chamfer} to distinguish changed or unchanged samples as below,
  \begin{align}
    d_{chamfer}(\bm{s}_1, \bm{s}_2)=\sum(\bm{s}_1\cdot\textrm{DT}(\bm{s}_2)+\bm{s}_2\cdot\textrm{DT}(\bm{s}_1))\label{eq:chamfer-distance}
  \end{align}
  where $s_{1}$, $s_{2}$ are images with the same size, $\textrm{DT}$ (Distance Transform) is an image expressing the distance to the nearest white pixel at each pixel, and the unit of $d_{chamfer}$ is [px].
  When $d_{chamfer}(\bm{s}_{start}, \bm{s}_{end}) \geq d_{thre}$, it is classified as a changed sample.
  For each tool, we collect $C_{changed}$ changed samples and $C_{unchanged}$ unchanged samples into a dataset ($C_{\{changed, unchanged\}}$ is the number of samples).
  After obtaining all these data, we go back to procedure (a) with a new tool.
  By repeating these procedures, we finally construct a dataset for the training of Tool-Net.

  When constructing a dataset, we augment it at the same time.
  When executing the procedures (c) -- (g), we collect not only ($\bm{s}_{start}$, $\bm{s}_{end}$, $\bm{t}$, $(\bm{\theta}^{T}_{start}, \bm{\theta}^{T}_{end})^{T}$) but also $\bm{s}$ and $\bm{\theta}$ at all frames.
  Then, we randomly choose frame indices $F_{\{from, to\}}$ ($F_{from}<F_{to}$) from the consecutively collected data, and add $C_{seq}$ data ($\bm{s}_{from}$, $\bm{s}_{to}$, $\bm{t}$, $(\bm{\theta}^{T}_{from}, \bm{\theta}^{T}_{to})^{T}$) into the dataset ($C_{seq}$ is the number of the data).
  $\{\bm{s}, \bm{\theta}\}_{\{from, to\}}$ is $\bm{s}$ or $\bm{\theta}$ at the frame of $F_{from}$ or $F_{to}$.
  When defining the arrangement of the robot and task image as shown in \figref{figure:robot-config}, we can also generate a mirrored data of ($\textrm{Mirror}(\bm{s}_{\{start, from\}})$, $\textrm{Mirror}(\bm{s}_{\{end, to\}})$, $\textrm{Mirror}(\bm{t})$, $-\bm{u}$) ($\textrm{Mirror}$ expresses a mirrored image).
  Therefore, we finally obtain $2(1+C_{seq})$ data: ($\bm{s}_{initial}$, $\bm{s}_{final}$, $\bm{t}$, $\bm{u}$).

  In our experiments, we set $\theta_{min}=-45$ [deg], $\theta_{max}=45$ [deg] regarding each actuator, $\theta_{thre}=45$ [deg], $d_{chamfer}=70.0$ [px], $C_{changed}=10$, $C_{unchanged}=5$, and $C_{seq}=24$.
}%
{%
  Tool-Netの学習に必要な$\bm{s}_{initial}$, $\bm{s}_{final}$, $\bm{t}$, $\bm{u}$を集める手順を以下に示す.
  \renewcommand{\labelenumi}{(\arabic{enumi})}
  \begin{enumerate}
    \item ロボットの姿勢を初期化し, 道具をロボットに取り付ける.
    \item 道具画像$\bm{t}$を取得する.
    \item $\bm{\theta}_{start}$をランダムに設定し, ロボットを動かす.
    \item ランダムに物体を配置する.
    \item タスク画像$\bm{s}_{start}$を取得する.
    \item $\bm{\theta}_{end}$をランダムに設定し, ロボットを動かす.
    \item タスク画像$\bm{s}_{end}$を取得する.
    \item (3) -- (7)を繰り返してデータを蓄積していき, 条件を満たしたら(1)へ戻る.
  \end{enumerate}

  (1)では, ロボットの姿勢を$\bm{0}$に初期化する.
  ロボットの関節角度を\figref{figure:robot-config}のように定義すると, これはつまり, ロボットアームを真っ直ぐにした状態のことを指す.
  また, 道具をロボットに取り付けるが, この道具は様々な形状を用意しなければならない.
  そこで, 本研究では\secref{subsec:experimental-setup}で説明するように道具に針金を用いるため, \figref{figure:random-tool}のようにランダムに制御点を配置し道具長さを一定にした3次Bスプラインを出力し, それを見ながら人間が手で道具を製作する.
  実際には針金に限らず, 3Dプリンタや粘土で製作しても良い.

  (3)と(6)では, ランダムにロボットを動作させるが, その際には$\bm{\theta}_{min} \leq \bm{\theta}_{\{start, end\}} \leq \bm{\theta}_{max}$という制限を設ける.
  ここで, $\bm{\theta}_{min}$, $\bm{\theta}_{max}$は関節角度の上下限を表す.
  また,  $\bm{\theta}_{start}$と$\bm{\theta}_{end}$が近すぎるとロボットがほとんど動作しないため, ランダムにそれらを選ぶ際に, $||\bm{\theta}_{start}-\bm{\theta}_{end}||_{1}>\theta_{thre}$という制限をつけている.
  ここで, $||\cdot||_{1}$はL1ノルムを表す.

  (8)では, データの偏りを減らすために, $\bm{s}_{start}$と$\bm{s}_{end}$の類似度を使って, 必要なデータのみを蓄積している.
  画像間の類似度(距離)には, 以下のようなsymmetric chamfer distance $d_{chamfer}$を用いる\cite{borgefors1988chamfer}.
  \begin{align}
    d_{chamfer}(\bm{s}_1, \bm{s}_2)=\sum(\bm{s}_1\cdot\textrm{DT}(\bm{s}_2)+\bm{s}_2\cdot\textrm{DT}(\bm{s}_1))\label{eq:chamfer-distance}
  \end{align}
  ここで, $s_{1}$, $s_{2}$は大きさの同じ二値画像, $\textrm{DT}$ (Distance Transform)は各画素値に対して直近の物体画像までの距離を表す画像を表し, $d_{chamfer}$の単位はpxである.
  $d_{chamfer}(\bm{s}_{start}, \bm{s}_{end}) < d_{thre}$のとき, タスク状態は変化していないとする.
  一つの道具に対して, タスク状態が変化したデータを$C_{changed}$個, 変化していないデータを$C_{unchanged}$個, データセットとして蓄積する.
  それらが全て取得し終わったら(1)に戻り, 新しい道具を取り付ける.
  これを繰り返すことでTool-Netのためのデータセットを作成していく.

  データ作成の際, データ拡張も同時に行う.
  (3) -- (7)の手順の際, ($\bm{s}_{start}$, $\bm{s}_{end}$, $\bm{t}$, $(\bm{\theta}^{T}_{start}, \bm{\theta}^{T}_{end})^{T}$)だけでなく, 常にタスク状態$\bm{s}$, 関節角度$\bm{\theta}$を蓄積しておく.
  そして, その連続したデータの中で, フレーム$F_{from}$, $F_{to}$ ($F_{from}<F_{to}$)をランダムに選び, ($\bm{s}_{from}$, $\bm{s}_{to}$, $\bm{t}$, $(\bm{\theta}^{T}_{from}, \bm{\theta}^{T}_{to})^{T}$)というデータを$C_{seq}$個データセットとして蓄積する.
  ここで, $\{\bm{s}, \bm{\theta}\}_{\{from, to\}}$はそれぞれフレーム$F_{from}$または$F_{to}$における$\bm{s}$または$\bm{\theta}$である.
  また, ロボットとタスク画像の位置関係が\figref{figure:robot-config}のように定義される場合は, $\bm{s}$, $\bm{t}$, $\bm{u}$をそれぞれ左右反転させてたデータも作成することができる.
  つまり, これまで得られた$1+C_{seq}$個のデータ$(\bm{s}_{\{start, from\}}, \bm{s}_{\{end, to\}}\bm{t}, \bm{u})$を, ($\textrm{Mirror}(\bm{s}_{\{start, from\}})$, $\textrm{Mirror}(\bm{s}_{\{end, to\}})$, $\textrm{Mirror}(\bm{t})$, $-\bm{u}$)として, 2倍に拡張することができる.
  ここで, $\textrm{Mirror}$は鏡面反転を表す.
  ゆえに, $2(1+C_{seq})$個のデータ($\bm{s}_{initial}$, $\bm{s}_{final}$, $\bm{t}$, $\bm{u}$)を得ることができる.

  本研究では, $\theta_{min}=-45$ [deg], $\theta_{max}=45$ [deg] (それぞれのモータに対して), $\theta_{thre}=45$ [deg], $d_{chamfer}=70.0$ [px], $C_{changed}=10$, $C_{unchanged}=5$, $C_{seq}=24$とする.
}%

\subsection{Training Phase of Tool-Net} \label{subsec:train-phase}
\switchlanguage%
{%
  We preprocess the obtained dataset ($\bm{s}_{initial}$, $\bm{s}_{final}$, $\bm{t}$, $\bm{u}$) and train Tool-Net.
  This preprocess makes the training result robust against noise and displacement of pixels.

  First, we augment tool image data $\bm{t}$ by adding noise.
  We show how to add noise in \algoref{algorithm:tool-noise}.
  \begin{algorithm}[t]
  \small
    \caption{Add noise to tool image $\bm{t}$}
    \label{algorithm:tool-noise}
    \begin{algorithmic}[1]
      \Function{AddNoiseToTool}{$\bm{t}$}
        \State $\bm{t}' \gets \bm{t}$
        \State $c_{add} \gets 0$
        \While{$c_{add} < C^{noise}_{add}$}
          \State $p = \textrm{ChooseRandomPixel}(\bm{t})$
          \State $c_{adjacent} = \textrm{CountAdjacentWhite}(\bm{t}, p)$
          \If{$\textrm{IsBlack}(\bm{t}, p)$ \AND $c_{adjacent}>0$}
            \State $\textrm{ToWhite}(\bm{t}', p)$
            \State $c_{add} = c_{add} + 1$
          \EndIf
        \EndWhile
        \While{$c_{del} < C^{noise}_{del}$}
          \State $p = \textrm{ChooseRandomPixel}(\bm{t})$
          \If{$\textrm{IsWhite}(\bm{t}, p)$}
            \State $\textrm{ToBlack}(\bm{t}', p)$
            \State $c_{del} = c_{del} + 1$
          \EndIf
        \EndWhile
        \State \Return $\bm{t}'$
      \EndFunction
    \end{algorithmic}
  \normalsize
  \end{algorithm}
  In \algoref{algorithm:tool-noise}, $\textrm{ChooseRandomPixel}(image)$ is the function which randomly extracts one pixel from $image$.
  $\textrm{CountAdjacentWhite}(image, pixel)$ is the function which counts the number of white pixels among 4 adjacent pixels to the $pixel$ of $image$.
  $\textrm{\{ToWhite, ToBlack\}}(image, pixel)$ is the function which makes the $pixel$ of $image$ \{white, black\}.
  $\textrm{\{IsWhite, IsBlack\}}(image, pixel)$ is the function which judges whether the $pixel$ in $image$ is \{white, black\}.
  $\bm{t}'$ is $\bm{t}$ with noise.
  $C^{noise}_{add}$, $C^{noise}_{del}$ are constant values.
  Line 4--11 of \algoref{algorithm:tool-noise} express that we randomly make the black pixel white when its adjacent pixels include at least one white pixel.
  Line 12--18 of \algoref{algorithm:tool-noise} express that we randomly make the white pixel black.

  Second, to make Tool-Net robust against the displacement of pixels, we blur the task state image $\bm{s}_{final}$, as shown below,
  \begin{align}
    \bm{s}'_{final} &= 1.0-\textrm{tanh}(C_{blur}\cdot\textrm{DT}(1-\bm{s}_{final})) \label{eq:blur}
  \end{align}
  where $C_{blur}$ is a constant value.
  The smaller $C_{blur}$ is, the more the image is blurred.

  Using $\bm{s}'_{final}$ and $\bm{t}'$, we train Tool-Net with the loss $L$ shown below, by setting the number of epochs as $C_{epoch}$ and batch size as $C^{train}_{batch}$,
  \begin{align}
    \bm{s}_{predicted} &= \bm{f}(\bm{s}_{initial}, \bm{t}', \bm{u})\\
    L &= \textrm{MSE}(\bm{s}_{predicted}, \bm{s}'_{final}) \label{eq:loss}
  \end{align}
  where $\textrm{MSE}$ expresses mean squared error.

  In the following experiments, we set $C_{blur}=0.2$, $C^{noise}_{add}=30$, $C^{noise}_{del}=30$, $C^{train}_{batch}=100$, and $C_{epoch}=300$.
}%
{%
  得られたデータセット($\bm{s}_{initial}$, $\bm{s}_{final}$, $\bm{t}$, $\bm{u}$)に処理を施しながら, Tool-Netを学習させる.

  まず, 学習の際の損失関数$L$は以下のように二乗誤差平均($\textrm{MSE}$)を用いる.
  \begin{align}
    \bm{s}_{predicted} = \bm{f}(\bm{s}_{initial}, \bm{t}, \bm{u}) \nonumber\\
    L = \textrm{MSE}(\bm{s}_{predicted}, \bm{s}_{final}) \label{eq:loss}
  \end{align}
  しかし, この損失関数では境界面が不連続になるためはpixelのズレに敏感であり学習が難しい.
  そこで, $\bm{s}_{final}$は以下のようにぼかして用いる.
  \begin{align}
    \bm{s}'_{final} &= 1.0-\textrm{tanh}(C_{blur}\cdot\textrm{DT}(1-\bm{s}_{final}))
  \end{align}
  ここで, $C_{blur}$は係数であり, 小さいほど画像がぼかされる.

  また, $\bm{t}$にはノイズを混入させて, データを拡張して用いる.
  この際のノイズの加え方を\algoref{algorithm:tool-noise}に示す.
  \begin{algorithm}[t]
    \caption{Add noise to tool image $\bm{t}$}
    \label{algorithm:tool-noise}
    \begin{algorithmic}[1]
      \Function{AddNoiseToTool}{$\bm{t}$}
        \State $\bm{t}' \gets \bm{t}$
        \State $c_{add} \gets 0$
        \While{$c_{add} < C^{noise}_{add}$}
          \State $p = \textrm{ChooseRandomPixel}(\bm{t})$
          \State $c_{adjacent} = \textrm{CountAdjacentWhite}(\bm{t}, p)$
          \If{$\textrm{IsBlack}(\bm{t}, p)$ \AND $c_{adjacent}>0$}
            \State $\textrm{ToWhite}(\bm{t}', p)$
            \State $c_{add} = c_{add} + 1$
          \EndIf
        \EndWhile
        \While{$c_{del} < C^{noise}_{del}$}
          \State $p = \textrm{ChooseRandomPixel}(\bm{t})$
          \If{$\textrm{IsWhite}(\bm{t}, p)$}
            \State $\textrm{ToBlack}(\bm{t}', p)$
            \State $c_{del} = c_{del} + 1$
          \EndIf
        \EndWhile
        \State \Return $\bm{t}'$
      \EndFunction
    \end{algorithmic}
  \end{algorithm}
  ここで, $\textrm{ChooseRandomPixel}(image)$は$image$からランダムに一つピクセルを抜き出す関数, $\textrm{CountAdjacentWhite}(image, pixel)$は$image$の$pixel$に隣接する4つの画素の中で白の画素の数を数える関数, $\textrm{To\{White, Black\}Pixel}(image, pixel)$は$image$の$pixel$を\{white, black\}にする関数, $\textrm{Is\{White, Black\}}(image, pixel)$は$image$の$pixel$が\{white, black\}かどうかを判断する関数である.
  また, $C^{noise}_{add}$, $C^{noise}_{del}$は定数である.
  これは, 隣接するピクセルに白の画素が一つでも含まれれば着目するピクセルを白にしても良い, ということを表す.

  この$\bm{s}'_{final}$と$\bm{t}'$ (ノイズを加えた$\bm{t}$)を用いて, \equref{eq:loss}のようにバッチサイズを$C^{train}_{batch}$, エポックを$C_{epoch}$として学習を行う.

  本研究では, $C_{blur}=0.2$, $C^{noise}_{add}=30$, $C^{noise}_{del}=30$, $C^{train}_{batch}=100$, $C_{epoch}=300$とする.
}%

\subsection{Optimization Phase of Tool-Net} \label{subsec:optimization-phase}
\switchlanguage%
{%
  We will explain the optimization procedures of tool shape and trajectory using trained Tool-Net.
  The procedures corresponding to \figref{figure:network-structure} are as below.
  \renewcommand{\labelenumi}{(\alph{enumi})}
  \begin{enumerate}
    \item Obtain the current task state $\bm{s}_{current}$ and target task state $\bm{s}_{target}$
    \item Generate the initial tool shape and trajectory $\{\bm{t}, \bm{u}\}_{init}$ before optimization
    \item Calculate loss of $\textrm{MSE}(\bm{f}(\bm{s}_{current}, \bm{t}_{init}, \bm{u}_{init}), \bm{s}'_{target})$
    \item Update $\{\bm{t}, \bm{u}\}_{init}$ through backpropagation
    \item Repeat (c) and (d) $C_{iter}$ times
  \end{enumerate}

  In (b), regarding $\bm{t}_{init}$, we randomly extract $C^{optimize}_{batch}$ tools from the dataset constructed in \secref{subsec:data-collection} ($C^{optimize}_{batch}$ is the number of the batch), and apply $\textrm{AddNoiseToTool}$ in \algoref{algorithm:tool-noise} to each tool.
  Regarding $\bm{u}_{init}$, we randomly generate $C^{optimize}_{batch}$ tool trajectories fulfilling the limit of $\bm{u}_{min} \leq \bm{u} \leq \bm{u}_{max}$.
  Thus, we construct a batch with $C^{optimize}_{batch}$ samples of $\bm{t}_{init}$ and $\bm{u}_{init}$.

  In (c), we calculate $L=\textrm{MSE}(\bm{f}(\bm{s}_{current}, \bm{t}_{init}, \bm{u}_{init}), \bm{s}'_{target})$ regarding each data in the batch ($\bm{s}'_{target}$ is $\bm{s}_{target}$ blurred by \equref{eq:blur}).

  In (d), we backpropagate $L$ and optimize $\bm{t}_{init}$ and $\bm{u}_{init}$ for each data.
  First, regarding the optimization of tool trajectory, we optimize $\bm{u}_{init}$ like in \cite{kawaharazuka2019dynamic, kawaharazuka2019flexible}, as below,
  \begin{align}
    \bm{g}_{control} &= dL/d\bm{u}_{init}\nonumber\\
    \bm{u}_{init} &\gets \bm{u}_{init}-\gamma\bm{g}_{control}/||\bm{g}_{control}||_{2} \label{eq:traj-opt}
  \end{align}
  where $||\cdot||_{2}$ expresses L2 norm, and $\gamma$ is an update rate.

  Second, regarding the optimization of tool shape, we change the values of pixels according to the gradient $\bm{g}_{tool} = dL/d\bm{t}_{init}$.
  To decrease $L$, the black pixels with negative gradients should be changed to white, and the white pixels with positive gradients should be changed to black.
  However, if all pixels are changed according to the gradient, an appropriate image for tool shape cannot be obtained due to sporadic pixels.
  Also, white pixels sometimes concentrate in small areas.
  To solve these problems, we optimize tool shape by focusing on adjacent pixels, as shown in \algoref{algorithm:tool-optimize}.
  \begin{algorithm}[t]
  \small
    \caption{Optimize tool image $\bm{t}$}
    \label{algorithm:tool-optimize}
    \begin{algorithmic}[1]
      \Function{CalcScore}{$\bm{t}, \bm{g}, p$}
        \State $value \gets 0$
        \State $c_{adj} = \textrm{CountAdjacentWhite}(\bm{t}, p)$
        \If{$\textrm{IsBlack}(\bm{t}, p)$ \AND $\textrm{IsNegGrad}(\bm{g}, p)$ \AND $c_{adj} > 0$}
          \State $value = \textrm{GetGrad}(\bm{g}, p)$
        \ElsIf{$\textrm{IsWhite}(\bm{t}, p)$ \AND $\textrm{IsPosGrad}(\bm{g}, p)$}
          \State $value = \textrm{GetGrad}(\bm{g}, p) + C_{scale}c_{adj}$
        \Else
          \State $value = C_{grad}\textrm{GetGrad}(\bm{g}, p)$
        \EndIf
        \State \Return $value$
      \EndFunction
      \Statex
      \Function{OptimizeTool}{$\bm{t}, \bm{g}$}
        \State $\bm{t}' \gets \bm{t}$
        \State $\bm{V} \gets []$
        \For{$p$ in $\bm{t}'$}
          \State $\textrm{push}(\textrm{CalcScore}(\bm{t}', \bm{g}, p), \bm{V})$
        \EndFor
        \State $P_{sorted} = \textrm{argsort}(\bm{V})$
        \For{$p$ in $\textrm{ExtractFront}(P_{sorted}, C^{optimize}_{add})$}
          \State $\textrm{ToWhite}(\bm{t}', p)$
        \EndFor
        \For{$p$ in $\textrm{ExtractBack}(P_{sorted}, C^{optimize}_{del})$}
          \State $\textrm{ToBlack}(\bm{t}', p)$
        \EndFor
        \State \Return $\bm{t}'$
      \EndFunction
    \end{algorithmic}
  \normalsize
  \end{algorithm}
  In \algoref{algorithm:tool-optimize}, $\textrm{\{IsPos, IsNeg\}Grad}(grad, pixel)$ is the function which judges whether the gradient $grad$ of $pixel$ is \{positive, negative\}.
  $\textrm{GetGrad}(grad, pixel)$ is the function which extracts $grad$ of $pixel$.
  $\textrm{\{ExtractFront, ExtractBack\}}(array, count)$ is the function which extracts $count$ values in $array$ from \{front, back\} of the array.
  $C_{scale}$, $C_{grad}$, $C^{optimize}_{add}$, and $C^{optimize}_{del}$ are constant values.
  We calculate a score for each pixel from the gradient and the number of adjacent white pixels.
  In ascending order of this score, the top $C_{add}$ and the bottom $C_{del}$ pixels are turned to white and black, respectively.
  $\bm{t}_{init}$ is updated by $\bm{t}_{init} \gets \textrm{OptimizeTool}(\bm{t}_{init}, \bm{g}_{tool})$ in \algoref{algorithm:tool-optimize}.
  % \algoref{algorithm:tool-optimize} makes black pixels, with more negative gradients and at least one white adjacent pixel, preferentially white.
  % Also, \algoref{algorithm:tool-optimize} makes white pixels, with the more positive gradients and more white adjacent pixels, preferentially black.

  After $C_{iter}$ iterations of \equref{eq:traj-opt} and $\textrm{OptimizeTool}(\bm{t}_{init}, \bm{g}_{tool})$, $C_{iter} \times C^{optimize}_{batch}$ candidates of tool shapes and trajectories are obtained.
  We use the tool shape and trajectory with minimum $L$ among all candidates as the optimized value of $\bm{t}_{optimized}$ and $\bm{u}_{optimized}$.

  We explained the method of optimizing both tool shape and trajectory at the same time.
  In the case that either tool shape or trajectory is optimized, the other is fixed and not optimized.
  In the following experiments, we set $C^{optimize}_{batch}=10$, $C_{iter}=50$, $\gamma = 0.1$ [rad], $C_{scale}=1E-3$, $C_{grad}=0.1$, $C^{optimize}_{add}=10$, and $C^{optimize}_{del}=10$.
}%
{%
  学習されたTool-Netを用いた道具形状・動作軌道の最適化の手順を以下に示す.
  本節では道具形状・動作軌道を同時に最適化する方法について述べるが, 道具形状のみ, または動作軌道のみの最適化は, 他方を任意の値で固定して最適化を行わないことで可能となる.
  その手順は以下であり, \figref{figure:network-structure}に対応する.
  \renewcommand{\labelenumi}{(\alph{enumi})}
  \begin{enumerate}
    \item 現タスク状態$\bm{s}_{current}$, 指令タスク状態$\bm{s}_{target}$を取得する.
    \item 道具形状と動作軌道の初期解$\{\bm{t}, \bm{u}\}_{init}$を作成する.
    \item 損失関数$\textrm{MSE}(\bm{f}(\bm{s}_{current}, \bm{t}_{init}, \bm{u}_{init}), \bm{s}_{target})$を計算する.
    \item 誤差逆伝播により, $\{\bm{t}, \bm{u}\}_{init}$を更新する.
    \item (c)と(d)を$C_{iter}$回繰り返す.
  \end{enumerate}

  (b)ではまず, 道具$\bm{t}_{init}$に関しては, 訓練時等のデータからランダムに$C^{optimize}_{batch}$個の道具を取り出し, それぞれに\algoref{algorithm:tool-noise}の$\textrm{AddNoiseToTool}$を適用する.
  また, 動作軌道$\bm{u}_{init}$に関しては, $\bm{u}_{min} \leq \bm{u} \leq \bm{u}_{max}$の中でランダムに$C^{optimize}_{batch}$個バッチを作成する.

  (c)ではこれら$C^{optimize}_{batch}$個のデータそれぞれに対して損失関数$L$を計算する.

  (d)では(c)で求まった$L$を誤差逆伝播し, バッチそれぞれに関して$\bm{t}_{init}$, $\bm{u}_{init}$を最適化する.
  まず, 動作軌道の最適化は\cite{tanaka2018emd, kawaharazuka2019dynamic, kawaharazuka2019flexible}と同様, 以下のように行う.
  \begin{align}
    \bm{g}_{control} &= dL/d\bm{u}_{init}\nonumber\\
    \bm{u}_{init} &= \bm{u}_{init}-\gamma\bm{g}_{control}/||\bm{g}_{control}||_{2}
  \end{align}
  ここで, $||\cdot||_{2}$はL2ノルム, $\gamma$は更新率を表す.

  次に, 道具形状の最適化の方法を\algoref{algorithm:tool-optimize}に示す.
  \begin{algorithm}[t]
    \caption{Optimize tool image $\bm{t}$}
    \label{algorithm:tool-optimize}
    \begin{algorithmic}[1]
      \Function{CalcScore}{$\bm{t}, \bm{g}, p$}
        \State $value \gets 0$
        \State $c_{adjacent} = \textrm{CountAdjacentWhite}(\bm{t}, p)$
        \If{$\textrm{IsBlack}(\bm{t}, p)$ \AND $\textrm{IsMinusGrad}(\bm{g}, p)$ \AND $c_{adjacent} > 0$}
          \State $value = \textrm{GetGrad}(\bm{g}, p)$
        \ElsIf{$\textrm{IsWhite}(\bm{t}, p)$ \AND $\textrm{IsPlusGrad}(\bm{g}, p)$}
          \State $value = \textrm{GetGrad}(\bm{g}, p) + C_{scale}c_{adjacent}$
        \Else
          \State $value = C_{grad}\textrm{GetGrad}(\bm{g}, p)$
        \EndIf
        \State \Return $value$
      \EndFunction
      \Statex
      \Function{OptimizeTool}{$\bm{t}, \bm{g}$}
        \State $\bm{t}' \gets \bm{t}$
        \State $\bm{V} \gets []$
        \For{$p$ in $\bm{t}'$}
          \State $\textrm{push}(\textrm{CalcScore}(\bm{t}', \bm{g}, p), \bm{V})$
        \EndFor
        \State $P_{sorted} = \textrm{argsort}(\bm{V})$
        \For{$p$ in $\textrm{ExtractFront}(P_{sorted}, C^{optimize}_{add})$}
          \State $\textrm{ToWhite}(\bm{t}', p)$
        \EndFor
        \For{$p$ in $\textrm{ExtractBack}(P_{sorted}, C^{optimize}_{del})$}
          \State $\textrm{ToBlack}(\bm{t}', p)$
        \EndFor
        \State \Return $\bm{t}'$
      \EndFunction
    \end{algorithmic}
  \end{algorithm}
  $\bm{g}_{tool} = dL/d\bm{t}_{init}$として, $\textrm{OptimizeTool}(\bm{t}_{init}, \bm{g}_{tool})$によって$\bm{t}_{init}$を更新していく.
  ここで, $\textrm{Is\{Plus, Minus\}Grad}(grad, pixel)$は画素$pixel$に関する勾配$grad$の正負を判定する関数, $\textrm{GetGrad}(grad, pixel)$は画素pixelに関する勾配$grad$を取り出す関数, $\textrm{Extract\{Front, Back\}}(array, count)$は\{front, back\}から数えて$count$個の値を$array$から取り出す関数である.
  また, $C_{scale}$, $C_{grad}$, $C^{optimize}_{add}$, $C^{optimize}_{del}$は定数である.
  これはつまり, 勾配が負(画素値を1にしたい)かつ現状画素値が0の画素の中で, 勾配の絶対値が大きいかつ隣接する画素に1つでも1が含まれる画素を優先的に1にしている.
  また, 勾配が正(画素値を0にしたい)かつ現状画素値が1の画素の中で, 勾配の絶対値が大きいかつ隣接する画素に1が多い画素を優先的に0にしている.
  これは, なるべく道具を連続したピクセルの繋がりによって表現するため, かつ勾配の高い画素が一箇所に集中してしまうのを防ぐためである.

  $C_{iter}$後, 最終的に全バッチ全エポックの中で最も$L$が小さい道具形状または動作軌道を, 最終的な値$\bm{t}_{optimized}$, $\bm{u}_{optimized}$として用いる.

  本研究では, $C^{optimize}_{batch}=10$, $C_{iter}=50$, $\gamma = 0.1$ [rad], $C_{scale}=1E-3$, $C_{grad}=0.1$, $C^{optimize}_{add}=10$, $C^{optimize}_{del}=10$とする.
}%

\subsection{Detailed Implementation} \label{subsec:detailed-implementation}
\switchlanguage%
{%
  The image binarization procedures of $\bm{s}$ and $\bm{t}$ are Crop, Color Extraction, Closing, Opening, and Resize, in order.
  Color Extraction separates the input image into tool, object, and background images.
  To make this process easy, we use a silver tool and the object is colored red.
  Note that the robot arm is considered as a background.
  Crop is executed as in \figref{figure:robot-config}, and Resize converts the image to the size of 64$\times$64.

  The convolutional layers for $\bm{s}$ and $\bm{t}$ have the same structures.
  Each of them has 6 layers, and the number of each channel is 1 (input), 4, 8, 16, 32, and 64.
  Its kernel size is 3$\times$3, stride is 2$\times$2, padding is 1, and batch normalization \cite{ioffe2015batchnorm} is applied after each layer.
  $\bm{s}$ and $\bm{t}$ are compressed to a 128 dimensional vector by fully connected layers, it is concatenated with $\bm{u}$, and a $256+2n$ dimensional vector is generated ($n$ is the number of actuators of the robot).
  After that, the vector is fed into fully connected layers whose numbers of units are $256+2n$, 128, 128, 128, and 256.
  The deconvolutional layers have the same structure with the convolutional layers, but only the last deconvolutional layer does not include batch normalization.
  The activation function of the last deconvolutional layer is Sigmoid, and those functions of the other layers are ReLU.
}%
{%
  $\bm{s}$, $\bm{t}$の二値化画像処理の流れは, crop, color extraction, closing, opening, resizeの順番で行う.
  それぞれ異なる色の背景, 道具, 物体を用いることで, 色抽出を容易にしている.
  cropは\figref{figure:robot-config}のような形で行い, resizeにより最終的に64$\times$64の画像に変換される.

  ネットワークの詳細な構造について説明する.
  $\bm{s}$と$\bm{t}$の畳み込み層は全く同じ構造をしている.
  それぞれ6層構造となっており, チャネル数はそれぞれ1 (input), 4, 8, 16, 32, 64とし, 全層においてKernel Size: 3x3, Stride: 2x2, Padding: 1とし, 全畳み込みの後にBatch Normalization \cite{ioffe2015batchnorm}を行う.
  その後, $\bm{s}$と$\bm{t}$はそれぞれ全結合層を通して128次元まで圧縮され, 圧縮された$\bm{s}$, $\bm{t}$, $\bm{u}$を結合して256+2n次元とする(ここで$n$はロボットの自由度を表す).
  その後, 256+2n, 128, 128, 128, 256というユニット数で全結合を行う.
  逆畳み込み層は畳み込み層を逆にした構造を持ち, 最終層のみBatch Normalizationを含まない.
  また, 最終層の活性化関数はSigmoidであり, それ以外の層における活性化関数はReLUを用いる.
}%

\section{Experiments} \label{sec:experiments}
\switchlanguage%
{%
  We will explain our experiments using the actual robot: the training of Tool-Net, the optimization of tool shapes using Tool-Net, and evaluation of the optimized tools.
  Also, we will show an advanced application of tool shape optimization for multiple tasks.
}%
{%
  実験セットアップ, Tool-Netの訓練, Tool-Netを用いた道具形状の最適化, Tool-Netを用いた道具形状最適化の実機における評価を順に行う.
  最後に, Tool-Netを用いた複数タスクに対する道具形状最適化という応用を示す.
}%

\subsection{Experimental Setup} \label{subsec:experimental-setup}
\begin{figure}[t]
  \centering
  \includegraphics[width=0.5\columnwidth]{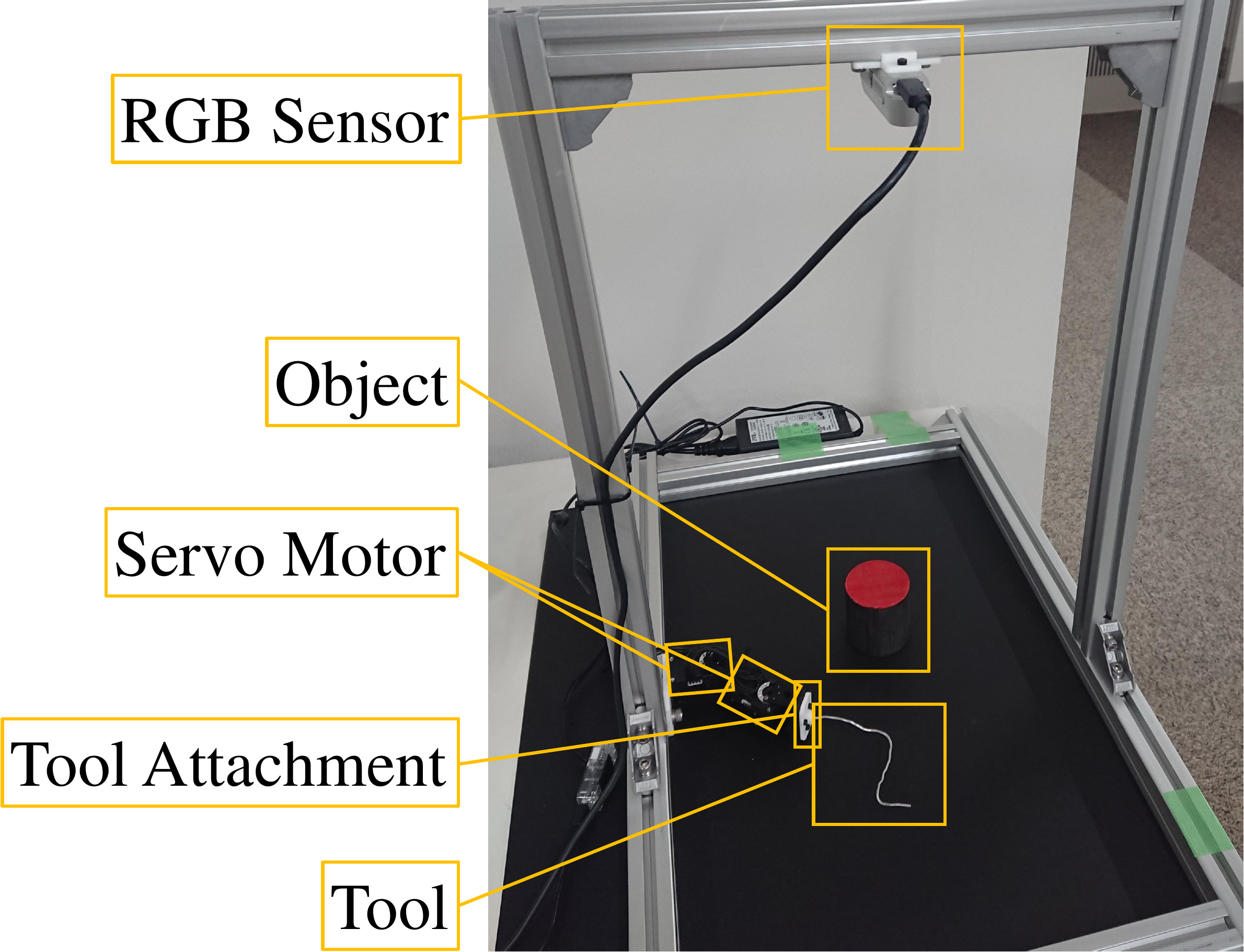}
  \caption{Experimental setup}
  \label{figure:experimental-setup}
  \vspace{-1.0ex}
\end{figure}

\begin{figure}[t]
  \centering
  \includegraphics[width=1.0\columnwidth]{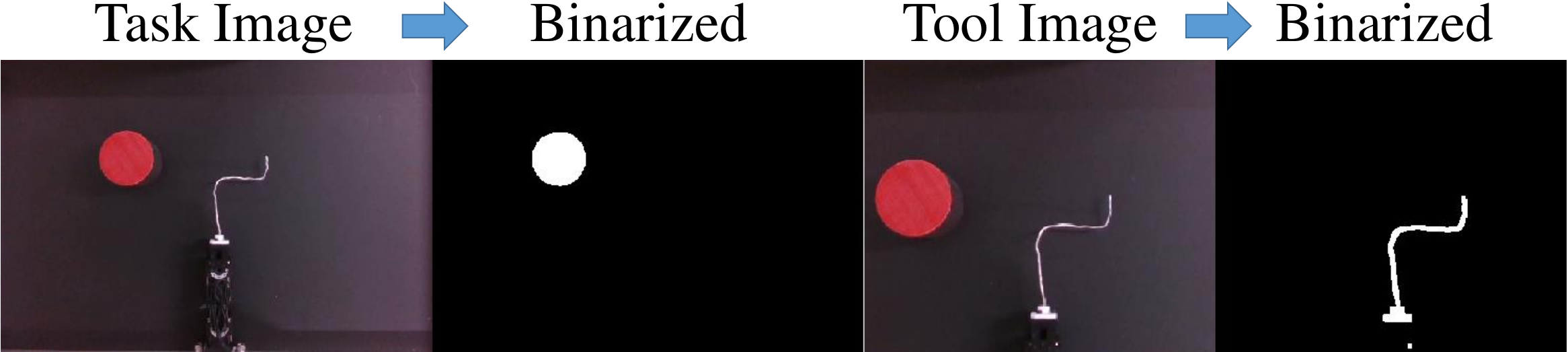}
  \vspace{-3.0ex}
  \caption{Results of image processing}
  \label{figure:process-experiment}
  \vspace{-3.0ex}
\end{figure}

\switchlanguage%
{%
  We show the experimental setup of this study in \figref{figure:experimental-setup}.
  Aluminum frames are structured on a black background sheet, and a camera and manipulator with 2 servo motors are attached to the structure.
  The servo motors are Dynamixel Motor (XM430-W350-R), and the camera is D435 (Intel Realsense).
  As a tool, we used metal wire with a diameter of 3 mm, which has enough strength for object manipulation tasks and can be bent by hand.
  A mount to attach the tool to is equipped at the tip of the manipulator.
  The object for the manipulation task is a cylinder-shaped wooden block painted red for color extraction.

  We show the images binarized by the method of \secref{subsec:detailed-implementation} in the left figure of \figref{figure:process-experiment}.
  The tool shape and task state are extracted and binarized as shown in the right figure of \figref{figure:process-experiment}, and we use them for Tool-Net after resizing.
}%
{%
  \figref{figure:experimental-setup}に本研究の実験セットアップを示す.
  黒い背景板の上にアルミフレームにより構成された支柱があり, 2自由度のマニピュレータとカメラが備え付けられている.
  MotorはDynamixel Motor (XM430-W350-R)を使用しており, RGB SensorとしてはD435 (Intel Realsense)を用いている.
  道具としては, タスクに対して十分な強度がある一方手で曲げることのできる太さの針金(直径3 mm)を用い, マニピュレータの先端には針金を刺して装着することができるようなマウントが備わっている.
  物体移動タスクに用いる物体は, 色抽出のために赤く塗られた円筒状の積み木である.

  D435から得られた画像を\secref{subsec:detailed-implementation}の方法で処理した結果を\figref{figure:process-experiment}に示す.
  このように道具とタスク画像を抽出・2値化しており, 最終的にはこれをresizeして用いる.
}%

\subsection{Tool Shape and Trajectory Optimization for One Task} \label{subsec:certain-experiment}
\subsubsection{Training Phase} \label{subsubsec:certain-experiment-training}
\begin{figure}[t]
  \centering
  \includegraphics[width=0.8\columnwidth]{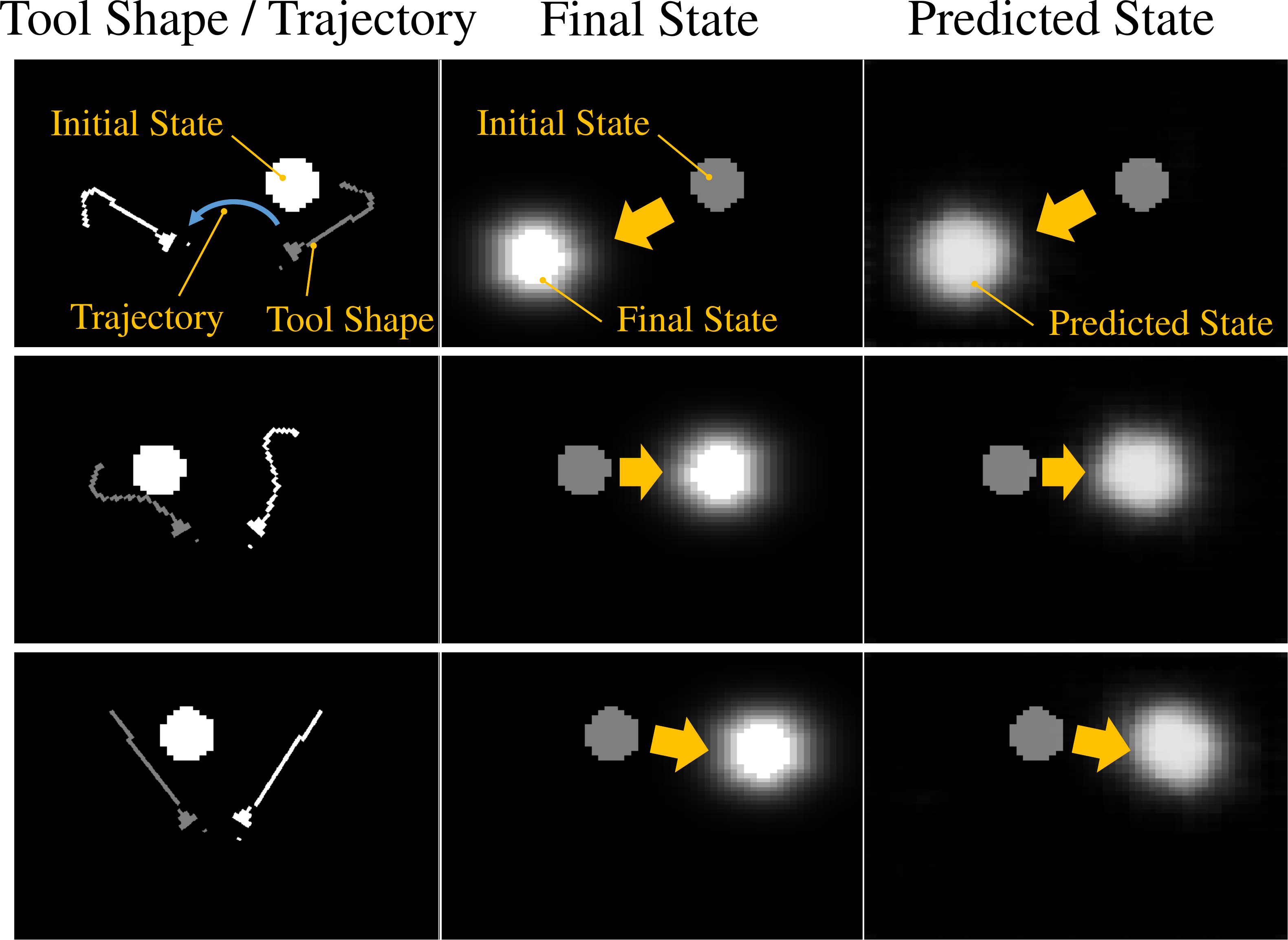}
  \caption{The inference results using the trained Tool-Net}
  \label{figure:predict-experiment}
  \vspace{-3.0ex}
\end{figure}

\switchlanguage%
{%
  We conducted data collection procedures for 48 kinds of tools over 2 hours in total, and obtained 36000 number of data.
  We trained Tool-Net using these data by $C^{train}_{epoch}$ epoch, and used the model with minimum $L$.
  We show the prediction results of task state in \figref{figure:predict-experiment}.
  When given a certain tool shape and trajectory, Tool-Net was able to predict the transition of task state correctly.
  We drew the tool trajectory by solving forward kinematics of the robot.
}%
{%
  約2hで道具48種類に対して実験を行い, データを合計36000個作成した.
  これを用いてTool-Netを学習させ, $C^{train}_{epoch}$回epochを回した際に最も$L$が小さかったモデルを使用する.
  このときのモデルを用いた指令タスク画像の予測結果を\figref{figure:predict-experiment}に示す.
  道具とその動きが与えられた際に, 正しくタスク状態の遷移を予測できていることがわかる.
  ToolのTrajectoryについては, ロボットの関節角度から順運動学を解いて画像を重ねて描画している.
}%

\subsubsection{Optimization Phase} \label{subsubsec:certain-experiment-optimization}
\begin{figure}[t]
  \centering
  \includegraphics[width=1.0\columnwidth]{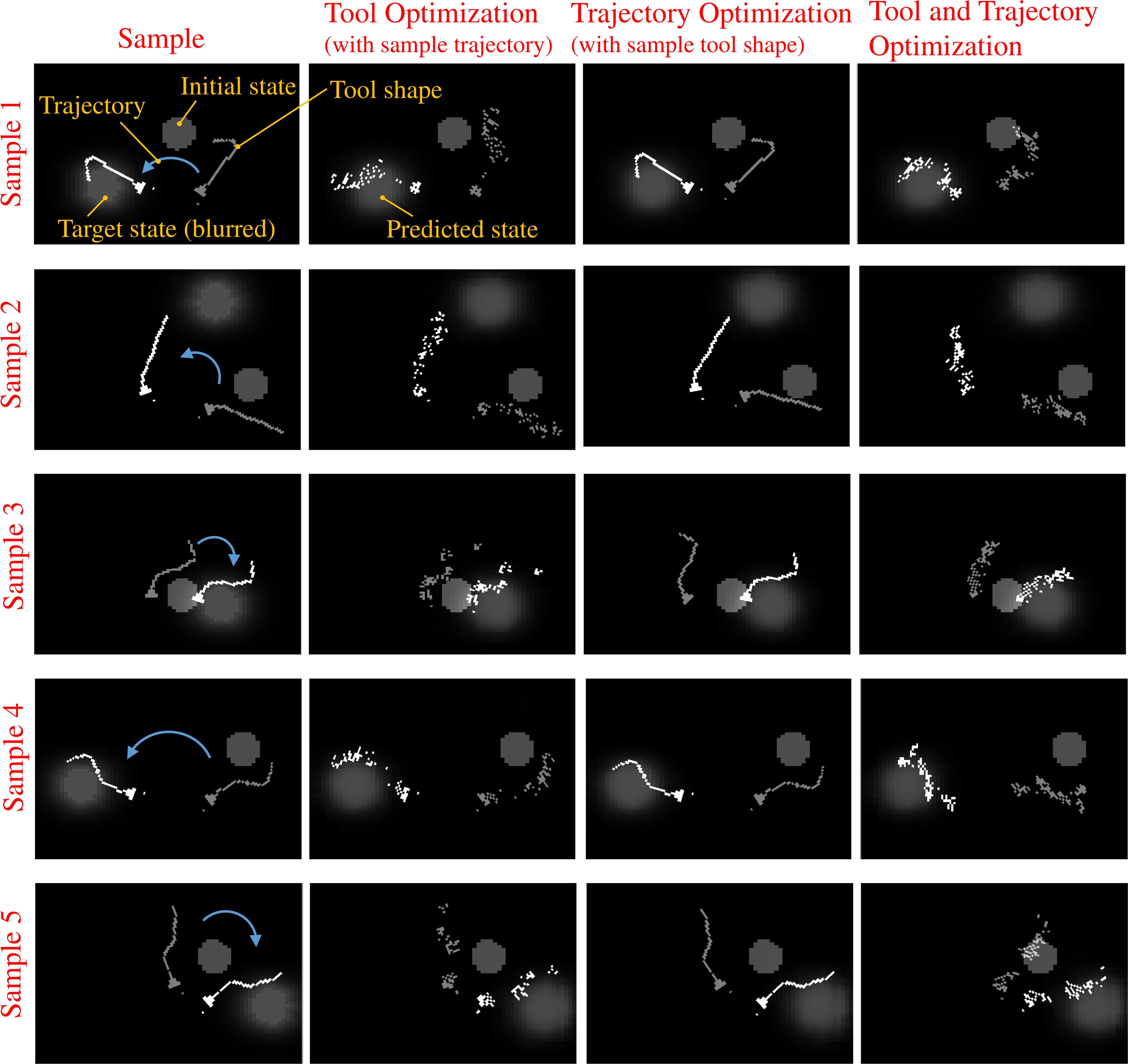}
  \vspace{-3.0ex}
  \caption{Samples of the optimized tool shape and trajectory using Tool-Net}
  \label{figure:optimize-experiment}
\end{figure}

\begin{figure}[t]
  \centering
  \includegraphics[width=1.0\columnwidth]{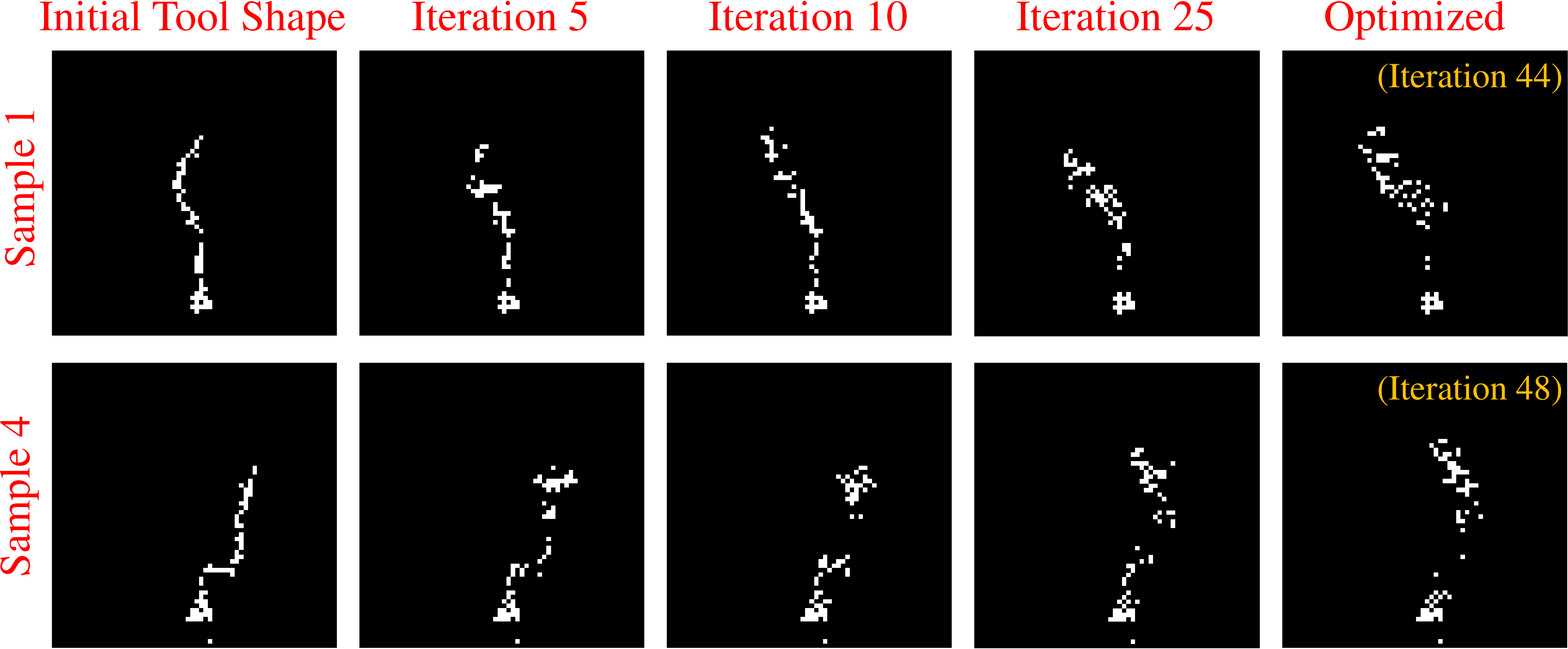}
  \vspace{-3.0ex}
  \caption{The transition of optimized tool shape by iterations, regarding Sample 1 and 4}
  \label{figure:optimize-transition}
  \vspace{-3.0ex}
\end{figure}

\switchlanguage%
{%
  We prepared Sample 1 -- 5 with the data of $(\bm{s}^{sample}_{initial}, \bm{s}^{sample}_{final}, \bm{t}^{sample}, \bm{u}^{sample})$.
  As we set $\bm{s}_{current}=\bm{s}^{sample}_{initial}$ and $\bm{s}_{target}=\bm{s}^{sample}_{final}$, we compared three optimizations: optimization of only tool shape, only tool trajectory, and both.
  When optimizing only tool shape, the tool trajectory is fixed to $\bm{u}^{sample}$, and when optimizing only tool trajectory, the tool shape is fixed to $\bm{t}^{sample}$.
  Note that the sample data is just for reference and $(\bm{t}^{sample}, \bm{u}^{sample})$ is not used for optimization except for the fixed trajectory or tool shape.
  We show the optimization results in \figref{figure:optimize-experiment}.
  % \figref{figure:optimize-experiment} shows the initial task state $\bm{s}^{sample}_{initial}$, $\bm{t}^{sample}$ or optimized tool shape, which is drawn using $\bm{u}^{sample}$ or optimized tool trajectory, and the predicted task state $\bm{s}^{sample}_{predicted}$ when using the tool shape and trajectory (the darknesses are adjusted).
  The left column shows sample data and the remaining three columns show the three optimization results.
  Although this optimization phase depends on the initial value, when $C^{optimize}_{batch}$ is large (e.g. $>1000$), almost the same results are obtained in every trial.

  Even though the tool shape is initialized by randomly chosen data from the training dataset and the tool trajectory is initialized by random values as explained in \secref{subsec:optimization-phase}, $\bm{s}^{sample}_{predicted}$ and $\bm{s}^{sample}_{final}$ indicate almost the same position and so the optimization succeeds.
  $\bm{s}^{sample}_{predicted}$ represents the predicted state when using the optimized tool shape and trajectory.
  For example, regarding Sample 1, a different tool shape from the sample data is generated, and it is reasonable for the task because it wraps the object well.
  Regarding Sample 2, a tool shape like the sample data is generated.
  Regarding the tool shape and trajectory optimization of Sample 3, a tool shape without the parts of the sample data, which do not contribute to the manipulation, is generated.
  We can see that various kinds of tool shapes are generated, not just the same shape with the sample data.

  Regarding the tool shape optimization of Sample 1 and 4, we show the transition of tool shape according to the iteration of optimization in \figref{figure:optimize-transition}.
  In actuality, although we start optimization from $C^{optimize}_{batch}$ initial tool shapes and choose the best one, we show only the transition of tool shape regarding the best tool finally chosen.
  % The size of tool shape image in \figref{figure:optimize-transition} is 64$\times$64, and the aspect ratio is different from in \figref{figure:process-experiment}.
  As we can see from the initial tool shape, the optimization starts from the tool shape that is close to the final shape, and it gradually changes.
}%
{%
  5つのサンプルデータSample (1) -- (5), $(\bm{s}^{sample}_{initial}, \bm{s}^{sample}_{final}, \bm{t}^{sample}, \bm{u}^{sample})$を用意する.
  それぞれの$(\bm{s}^{sample}_{initial}, \bm{s}^{sample}_{final})$に対して, 道具形状のみを最適化する場合, 動作軌道のみを最適化する場合, 道具形状と動作軌道を同時に最適化する場合に関して比較を行う.
  道具形状のみを最適化する場合は動作軌道は$\bm{u}^{sample}$を, 動作軌道のみを最適化する場合は道具形状は$\bm{t}^{sample}$を用いる.
  最適化結果をロボットの順運動学を解いて画像を重ねて示したものを\figref{figure:optimize-experiment}に示す.
  ここで表示しているのは, タスクの初期状態$\bm{s}^{sample}_{initial}$, $\bm{t}^{sample}$または最適化された道具の形状を, $\bm{u}^{sample}$または最適された動作軌道をもとに描画したもの, その道具と軌道によって予測されるタスク状態$\bm{s}^{sample}_{predicted}$である(色の濃さは適宜変えている).

  \secref{subsec:optimization-phase}で述べたように, 道具は訓練時データからランダムに取り出して最適化の初期値とし, 動作軌道はランダムな値を最適化の初期値としているが, 最適化の結果として, $\bm{s}^{test}_{predicted}$と$\bm{s}^{test}_{final}$がほぼ同じ場所を示しており, 最適化が成功していることがわかる.
  例えばSample (1)を見ると, 元のデータとは異なる道具が生成されており, 物体を包み込んで動かすような, タスク達成にもリーズナブルな結果となっていることがわかる.
  Sample (2)では元データと同じような道具が生成され, Sample (3)の道具形状と動作軌道の最適化では元の道具から無駄な部分を削ぎとったような道具形状が生成されている.
  このように, 様々な形態の道具が生成されていることがわかる.

  Sample (1), (4)における道具のみの最適化に関して, 最適化のイテレーションに伴う道具形状の変化の様子を\figref{figure:optimize-transition}に示す.
  実際には$C^{optimize}_{batch}$個の初期解からスタートし最も良い道具を選ぶが, \figref{figure:optimize-transition}には, 最終的に最も損失の小さかった道具の初期解からの遷移を表示している.
  また, 道具画像は 64$\times$64であるため, \figref{figure:process-experiment}等に示される本来のアスペクト比とは異なることに注意されたい.
  道具の初期解からわかるように, 多少最適化後に近い道具からスタートし, 徐々に最終的な道具へと変化していることがわかる.
}%

\subsubsection{Evaluation of Optimization Phase} \label{subsubsec:certain-experiment-eval}
\begin{figure}[t]
  \centering
  \includegraphics[width=1.0\columnwidth]{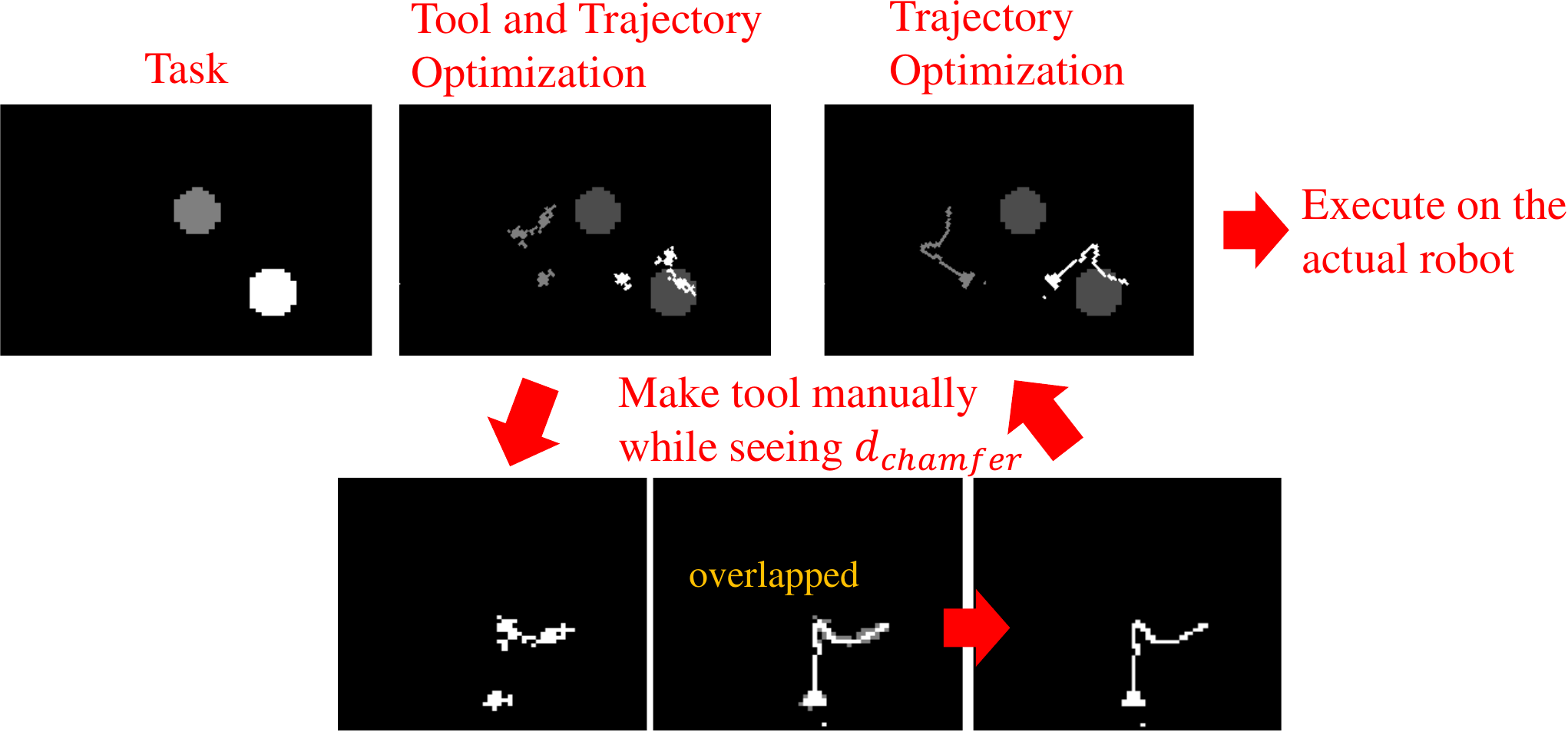}
  \vspace{-3.0ex}
  \caption{Procedures of task execution using optimized tool shape on the actual robot}
  \label{figure:optimize-procedure}
\end{figure}

\begin{figure}[t]
  \centering
  \includegraphics[width=1.0\columnwidth]{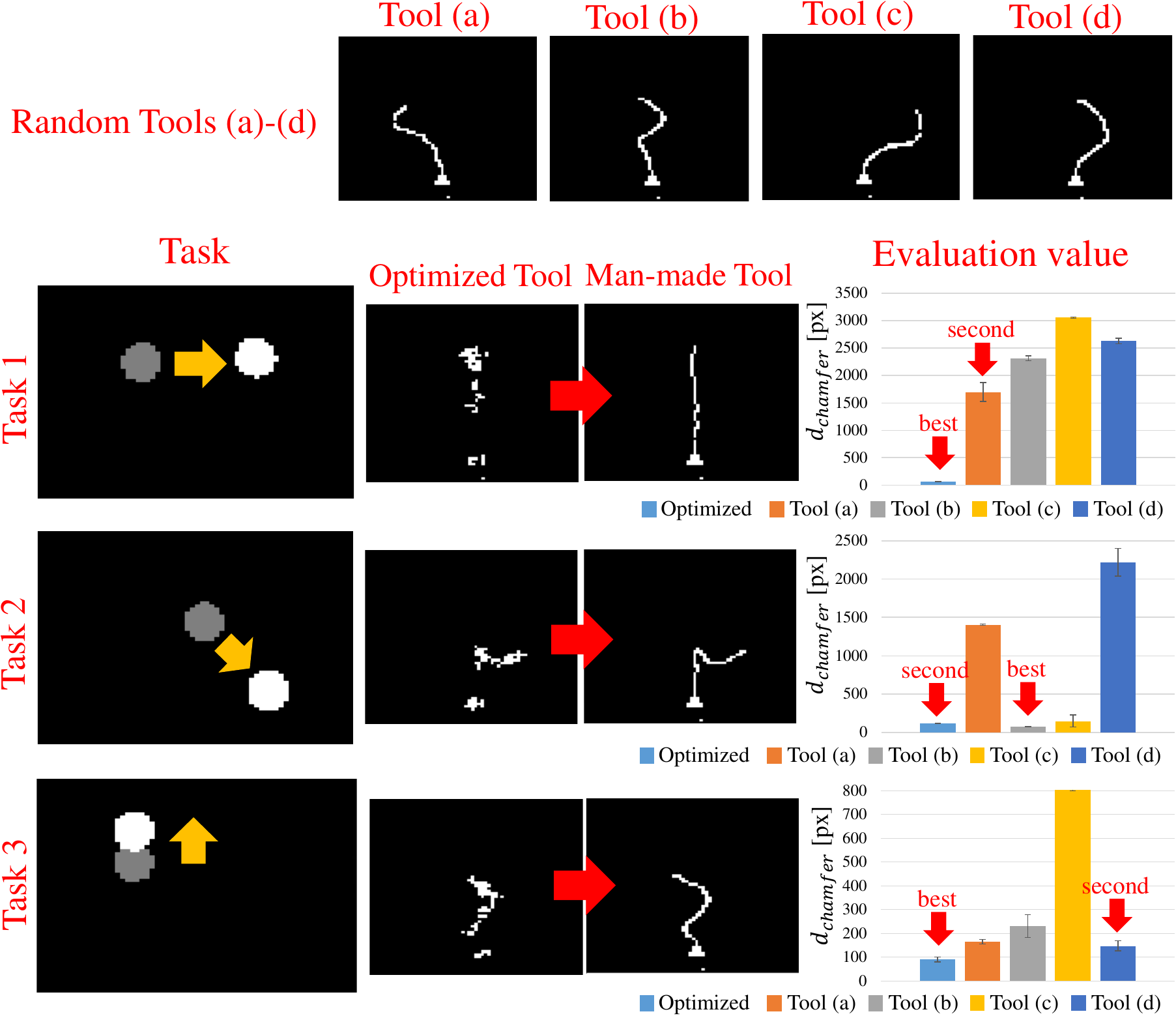}
  \vspace{-3.0ex}
  \caption{Evaluation of tool shape optimization}
  \label{figure:optimize-eval}
  \vspace{-3.0ex}
\end{figure}

\switchlanguage%
{%
  Regarding the tool shape and trajectory optimization, we evaluated the degrees of task realization in an actual robot experiment.
  We prepared Task 1 -- 3 with the data of $(\bm{s}^{test}_{current}, \bm{s}^{test}_{target})$.
  By the procedures shown in \figref{figure:optimize-procedure}, the task is executed with the optimized tool, and task realization is evaluated.
  First, the tool shape and trajectory are optimized at the same time for the given task.
  Second, the optimized tool shape is made with metal wire by human hands.
  When $d_{chamfer}$ between optimized and current tool shapes becomes lower than 150 px, it is regarded that the same tool shape is made.
  Although this procedure includes some human arbitrariness, the threshold was enough to resemble the generated tool in this experiment.
  Third, only the tool trajectory is optimized for the man-made tool, and the optimized motion is executed.
  After the task execution, $d_{chamfer}$ between the final task state and the target task state is measured.
  We repeated the procedures from task execution to measurement of $d_{chamfer}$ 5 times, and calculated its average and variance.
  Also, we prepared 4 random tool shapes: Tool (a) -- (d), for comparison as shown in the upper figure of \figref{figure:optimize-eval}.
  Regarding each random tool, only the tool trajectory is optimized, the optimized motion is executed, and the average and variance of $d_{chamfer}$ are calculated.

  Regarding each task, we show the task, optimized tool shape, man-made tool shape based on the optimized one, and the average and variance of $d_{chamfer}$, in \figref{figure:optimize-eval}.
  From the average of $d_{chamfer}$, we can say that the degrees of task realization depend on the tool shapes, and they are high in general when using the optimized tools.
  % Regarding Task 1, although the task cannot be achieved well when using Tool (a) -- (d) because the object is distant from the tool, the optimized tool shape is straight and the distant object is manipulated well.
  Regarding Task 1, in which the object is placed far from the robot, the optimized tool is straight-shaped and can achieve the task well, where Tool (a) -- (d) cannot reach the object.
  Regarding Task 2, the tool shapes of Tool (b), (c) and the optimized one have the same shape to wrap and pull the object to the right front side, and the $d_{chamfer}$ are almost the same.
}%
{%
  Tool-Netを用いた道具形状・動作軌道の最適化に関する, 実機におけるタスク実現度の評価を行う.
  まず, 訓練時とは異なる3つのタスクデータ Task (1) -- (3) $(\bm{s}^{test}_{current}, \bm{s}^{test}_{target})$を用意する.
  ここで道具最適化について, \figref{figure:optimize-procedure}のような形で実際にロボットにおいて実行し, タスク実現度を評価する.
  まず与えられたタスクに対して道具形状と動作軌道を同時に最適化する.
  そこで得られた道具形状に近くなるように, 人間が針金を曲げ道具形状を作成し, 最適化された道具形状と作成された道具形状の間の$d_{chamfer}$が150 px以下となったところで, 同様の道具が作成できたと見なす.
  そして, この作成された道具を用いて動作軌道のみを最適化し, $\bm{s}^{test}_{current}$の状態からその動作を実機において実行する.
  実機動作後のタスク状態と指令タスク状態$\bm{s}^{test}_{target}$の間の$d_{chamfer}$を測定する.
  動作実行から$d_{chamfer}$の測定までを5回行い, 平均と分散を計算する.
  また, 比較としてランダムな道具のデータを4つ用意し(\figref{figure:optimize-eval}の上図), それぞれの道具についてTool-Netを用いて動作軌道のみ最適化し, 実ロボットで動作を実行し, 同様に$d_{chamfer}$の平均と分散を計算する.

  タスク, 最適化された道具形状, 作成された道具形状, $d_{chamfer}$の平均と分散を\figref{figure:optimize-eval}の下図に示す.
  $d_{chamfer}$の平均から, 道具によってタスクの得意不得意があることがわかるが, 最適化後の道具を用いた場合は, 総じてタスクの実現度が高いことがわかる.
  Task (1)では, 物体が遠いためTool (1) -- (4)では上手く操作することができないのに対して, 最適化の結果として真っ直ぐな, より遠くの物体を扱える道具が生まれている.
  Task (2)では, Tool (2), Tool (3), Optimizedは同じように物体を右手前へ引くための鍵のような形をしており, タスク実現度もほぼ同じであることがわかる.
}%

\subsection{Tool Shape Optimization for Multiple Tasks} \label{subsec:multitask-experiment}
\begin{figure}[t]
  \centering
  \includegraphics[width=1.0\columnwidth]{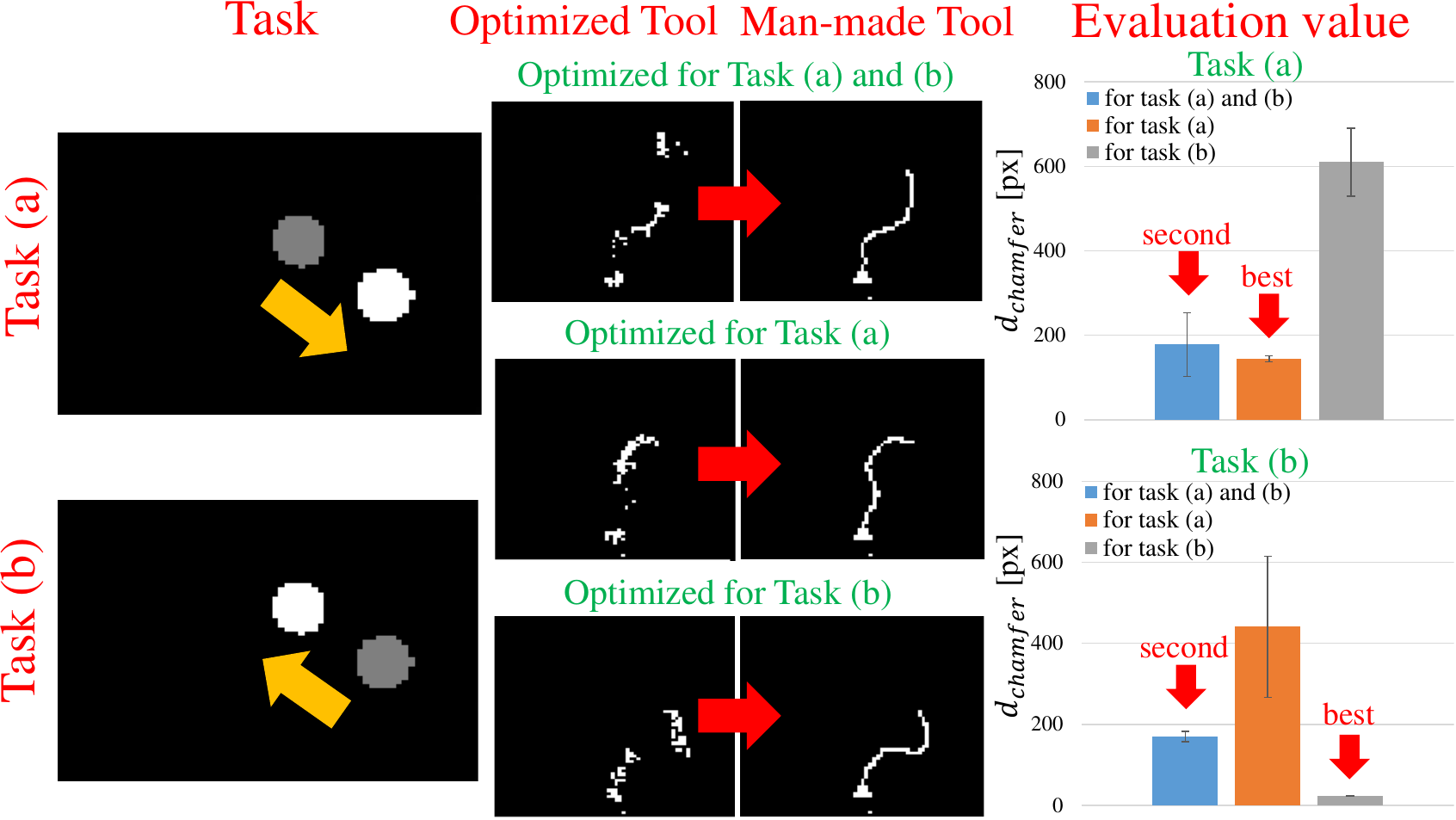}
  \vspace{-3.0ex}
  \caption{Experimental evaluation of tool shape optimization for multiple tasks}
  \label{figure:multitask-experiment}
  \vspace{-3.0ex}
\end{figure}

\switchlanguage%
{%
  The optimization of tool shape in this study has a potential to be applied for not only one task but also multiple tasks.
  This can be executed by summing up $L$ obtained for multiple tasks and backpropagating it.

  To evaluate the applicability for multiple tasks, we prepared reversed tasks, Task (a): $\bm{s}^{test}_{1}\to\bm{s}^{test}_{2}$ and Task (b): $\bm{s}^{test}_{2}\to\bm{s}^{test}_{1}$.
  We executed experiments of the tool shape and trajectory optimization for only Task (a), only Task (b), and both, like in \secref{subsubsec:certain-experiment-eval}, and then, calculated the average and variance of $d_{chamfer}$ between the final task state and the target task state.
  We show the results in \figref{figure:multitask-experiment}.
  The tool shape optimized for either Task (a) or (b) has a shape to wrap the object firmly.
  On the other hand, the tool shape optimized for both Task (a) and (b) has a shape with a gentle curve, and can be used for both Task (a) and (b).
  Regarding $d_{chamfer}$, when using the optimized tool shape for only Task (a), Task (a) is realized well, but Task (b) cannot be realized.
  On the other hand, when using the optimized tool shape for both Task (a) and (b), both tasks can be realized to some extent.
}%
{%
  本研究は一つのタスクに対する道具最適化のみならず, 複数のタスクに対する道具最適化を行うことが可能である.
  これは, 複数のタスクに対して得られた損失関数$L$の合計値を誤差逆伝播するのみである.
  まず, 異なる2つのタスクデータを用意するが, 本研究では\figref{figure:multitask-experiment}のように反転したTask (a): $\bm{s}^{test}_{1}\to\bm{s}^{test}_{2}$とTask (b): $\bm{s}^{test}_{2}\to\bm{s}^{test}_{1}$を用意する.
  ここで, 道具をTask (a)のみに対して最適化した場合, Task (b)のみに対して最適化した場合, Task (a) and (b)に関して最適化した場合について, \secref{subsubsec:certain-experiment-eval}と同様に実験を行い, $d_{chamfer}$の平均と分散を求める.
  その結果を\figref{figure:multitask-experiment}に示す.
  Task (a), Task(b)それぞれのみに対して最適化された道具は, 物体をしっかりと包むような形状の道具になっていることがわかる.
  これに対して, Task (a)と(b)に関して最適化された道具は傾斜が緩く, 両者に対して使えるようになっている.
  タスク実現度も, Task (a)のみに最適化された道具では, Task (a)の実現度は最も高いのに対して, Task (b)の実現度は著しく低く, Task (b)のみに最適化された道具も同様である.
  一方で, 両者に対して最適化された道具では, どちらのTaskもある程度正確にこなすことができていることがわかる.
}%

\section{Discussion} \label{sec:discussion}
\subsection{Experimental Results}
\switchlanguage%
{%
  The experimental results in \secref{subsubsec:certain-experiment-training} indicate that Tool-Net can infer the change of task state from the current task state, tool shape, and tool trajectory.
  \secref{subsubsec:certain-experiment-optimization} demonstrates that various tool shapes and tool trajectories are generated as a result of the optimization process.
  These tool shapes are usually reasonable, because they wrap the target object well, have no useless parts that do not contribute to the manipulation, etc.
  However, their pixels lose continuity.
  In \secref{subsubsec:certain-experiment-eval}, humans made the optimized tool shape by hand while referencing $d_{chamfer}$, and the optimized trajectory was executed.
  As a result, the optimized tool shapes can achieve the target tasks better than the randomly generated tool shapes.
  While the randomly generated tools usually cannot reach the target object or cannot wrap the object well, the optimized tools can reach and wrap it well.
  In \secref{subsec:multitask-experiment}, Tool-Net is also applicable to multiple tasks.
  While a tool optimized for a certain task can realize the task well, the tool shape is hard to be used for other tasks.
  In contrast, while a tool optimized for multiple tasks is a little inferior to the tools optimized for each task, the tool shape has versatility that can be used for the multiple tasks.

  Our system has a problem on the making of actual tools, because the generated tool shapes lose continuity of pixels and cannot be directly made.
  Because humans make a tool similar to the optimized one while referencing $d_{chamfer}$ in this study, this process depends on human interpretation.
  To solve the problem, we need to develop techniques of generating realizable tool shapes while keeping the diversity of tool shapes.
}%
{%
  \secref{subsubsec:certain-experiment-training}から, Tool-Netは現在タスク状態・道具形状・道具軌道を与えることで, タスク状態の変化を計算することができる.
  また, \secref{subsubsec:certain-experiment-optimization}の結果から, 最適化の結果として, 多様な道具形状・道具軌道が生成されることがわかった.
  それらは, 物体をしっかり包み込んだり, 無駄のない作りだったりとリーズナブルなものが多い一方, ピクセルの途切れによって良いかどうかを判断することが難しいものが多かった.
  \secref{subsubsec:certain-experiment-eval}では, 最適化された道具形状を人間が$d_{chamfer}$を見ながら生成し, 動作を実行させた.
  その結果, ランダムに生成した道具よりも, 最適化された道具はタスク実現に良い結果を残した.
  ランダムに生成した道具は, 物体が届かずにタスクを実行できなかったり, 物体を上手く捉えられずに途中でズレてしまったりしていたのに対して, 最適化された道具は物体を上手く捉えることができていた.
  また, \secref{subsec:multitask-experiment}では複数タスクへの適用が可能なことが示された.
  単体タスクに対する最適化をした場合はそのタスクを実現するうえで最も有利な道具が生成される一方, 他のタスクでは使いにくい道具が生成されてしまう.
  これに対して, 複数をタスクへの最適化を行えば, 単体タスクへの最適化を施した道具には及ばないものの, 複数のタスクである程度良い結果を残すことができる道具が生成されることを確認した.

  しかし問題点として, 先ほど説明したように, 生成された道具のピクセルが途切れ途切れになってしまうため, その道具を生成するのが難しいということが挙げられる.
  本研究では$d_{chamfer}$を見ることで似た形状の道具を人間が生成したが, これは人間の解釈に依存してしまうところがあるため望ましくない.
  これを解決するためには, 道具のピクセルが途切れないように最適化を施す技術, または, 得られた画像を上手く繋いで実現可能な道具画像に変換する技術が求められる.
}%

\subsection{Future Directions}
\switchlanguage%
{%
  We believe that this framework can be applied to not only object manipulation but also more general tasks, by changing the definition of task state.
  For example, by using 6 axis force sensor or frequency and amplitude of sound as the task state, a force applying or sound making task could be achieved.

  Because the metal wire is used for experiments, the generated tool shape is limited to the shape drawn by one stroke.
  This spoils the benefits of using the binarized image as tool expression.
  If we can use 3D printer or clay for tool-making, the benefits of using an image can be emphasized more, because tool shapes with branches and larger areas can be handled.

  Tool material or friction coefficient also sometimes affects task execution.
  Although a binarized image cannot express these characteristics, we may be able to optimize tool shape considering them by embedding this information into an image with multiple channels like color image.

  If we would like to handle 3D movement, by using 3D voxel representation with depth image for tool shape, we believe this study could be extended to the 3D movement.

  This framework could be also applicable to the flexible manipulator, if the representation of tool trajectory is changed by adding time information to $\bm{u}$ or making the network a recurrent one.
  Such networks will have a similar structure with \cite{kawaharazuka2019dynamic}.

  We came up with this study when seeing a human drop a key into a gutter by mistake and pick it up using metal wire.
  To achieve this task, not only the expansion to 3D movement, but also a consideration of obstacles and efficient learning will be required.
  In the current form, the necessary number of trials explodes exponentially depending on the number of robot actuators and degrees of freedom of tool shape representation.
  It will be important for efficient training of Tool-Net to address the curse of dimensionality by using not only obtained data but also prior knowledge such as physical laws and the analogy of own body and tool shape like in \cite{tee2018tool}.
}%
{%
  本研究は, タスク状態の定義を変更することで, 力を加えたり音を出したりするようなタスク等にも同様に適用可能である.
  力の場合には6軸力センサの値, 音の場合は周波数と振幅等がタスク状態として考えられる.

  本研究では道具形状を2次元の画像として表したが, 本実験においては道具として針金を使っているため, 一筆書きできる形状しか製作することはできない.
  粘土や3Dプリンタ, 道具同士の組み合わせを用いることで, 今後より道具形状として画像を使う利点が強調されると考える.
  一方, 二値画像では道具の素材・摩擦係数等に関する情報を得ることは難しく, カラー画像を用いることで材質も考慮した道具形状最適化が可能となる可能性がある.
  また道具に3次元のボクセル表現を用いることで, 3次元運動に拡張することも可能であると考える.

  他にも, 本フレームワークを柔軟マニピュレータに同様に適用することが可能である.
  その場合は, $\bm{u}$に時間を入れたり時系列にしたり等, 制御入力が動的要素を含む形に変更する必要がある.

  本研究は網のかかった側溝に鍵を落とした人間が, 針金を使って取り出すのを見て思いついた.
  これを成功させるためには, 先ほど述べた3次元への拡張だけでなく, 障害物を考慮すること, 学習の効率化等が必要になると考える.
  現状の動作軌道生成では, ロボットの自由度や道具の表現自由度が増えるに従い計算量が指数関数的に爆発する.
  全てを一から学習するのではなく, これまでの様々な物体操作や道具使用における知識, \cite{tee2018tool}のような自分の身体と道具形状の類似性等を取り込み, 学習の効率化を測ることが重要である.
}%

\section{CONCLUSION} \label{sec:conclusion}
\switchlanguage%
{%
  We proposed a method to obtain an optimized tool shape and trajectory for given tasks using backpropagation technique of a neural network.
  A transition network of task state by a certain tool shape and trajectory is trained, and a tool shape and trajectory to realize the target task state are calculated using it.
  Also, we proposed data augmentation for efficient training, and a method to update pixels in tool shape image for optimization.
  Finally, the given tasks can be achieved more accurately by using the optimized tool shape.
  In future works, by using not only obtained data but also prior knowledge such as physical laws and robot configuration, we will expand this method to a more practical form.
}%
{%
  本研究では, 与えられたタスクに対して, ニューラルネットワークの誤差逆伝播を用いて道具形状・動作軌道を最適化する手法を提案した.
  ある道具形状とその動作軌道によるタスク状態の変化を記述するネットワークを構築し, 指令タスク状態を実現するように道具形状・動作軌道を更新していく.
  また, 効率的な学習のためのデータ拡張方法, 画像で表現した道具の最適化時のピクセル更新手法等についても論じた.
  結果として, 道具最適化により, より正確にタスクを実現することが可能となった.
  今後は, データだけでなく物理法則や事前知識を統合し, 本手法をより実用的な形へと発展させていきたい.
}%

{
  %\footnotesize
  %\small
  %\bibliographystyle{junsrt}
  \bibliographystyle{IEEEtran}
  \bibliography{main}
}

\end{document}